%% file: main.tex
\documentclass[sigconf,balance=false,authorversion,nonacm]{acmart}
\usepackage{amsmath}
\usepackage{subcaption}
\usepackage{adjustbox}
\usepackage{nicefrac}
\usepackage{siunitx}
\usepackage{array,framed}
\usepackage{booktabs}
\usepackage{textcomp}
\usepackage{setspace}
\usepackage{hyperref}
\usepackage{float}
\usepackage{latexsym,fancyhdr,url}
\usepackage{enumerate}
\usepackage{algorithm}
\usepackage{algorithmicx}
\usepackage{algpseudocode}
\usepackage{graphics}
\usepackage{xparse}
\usepackage{xspace}
\usepackage{xcolor}
\usepackage{multirow}
\usepackage{csvsimple}
\usepackage{balance}
\usepackage{verbatim}
\usepackage{
  tikz,
  pgfplots,
  pgfplotstable
}

\usetikzlibrary{
  shapes.geometric,
  arrows,
  external,
  pgfplots.groupplots,
  matrix
}

\pgfplotsset{compat=1.9}
\usepackage{mathtools}
\usepackage{anyfontsize}
\newcommand{\bl}[1]{\textcolor{black}{#1}}

\begin{document}

\title[Exploiting XAI for Model Extraction Attacks against Interpretable Models]{AUTOLYCUS: Exploiting Explainable Artificial Intelligence (XAI) for Model Extraction Attacks against Interpretable Models}
\titlenote{Published in the Proceedings on Privacy Enhancing Technologies (PoPETs), Vol. 2024, Issue 4, 2024.}
\titlenote{Refer to \url{https://github.com/acoksuz/AUTOLYCUS} for the source code.}


\author{Abdullah Caglar Oksuz}
\orcid{0000-0001-8530-9960}
\affiliation{%
  \institution{Case Western Reserve University}
  \city{Cleveland}
  \state{Ohio}
  \country{USA}}
\email{abdullahcaglar.oksuz@case.edu}

\author{Anisa Halimi}
\affiliation{%
  \institution{IBM Research - Dublin}
  \city{Dublin}
  \country{Ireland}}
\email{anisa.halimi@ibm.com}

\author{Erman Ayday}
\affiliation{%
  \institution{Case Western Reserve University}
  \city{Cleveland}
  \state{Ohio}
  \country{USA}}
\email{erman.ayday@case.edu}

\renewcommand{\shortauthors}{Oksuz et al.}

\input{sections/abstract}


\maketitle

\input{sections/intro}    
\input{sections/related_work}
\input{sections/preliminaries}
\input{sections/methodology}
\input{sections/results}
\input{sections/countermeasures}
\input{sections/discussion}
\input{sections/conclusion}

\begin{acks}
For the research reported in this publication, Abdullah Caglar Oksuz and Erman Ayday was supported in part by the National Science Foundation (NSF) under grant number OAC-2112606. Anisa Halimi was partly supported by the European Union's Horizon 2020 research and innovation programme under grant number 951911 – AI4Media. The content is solely the responsibility of the authors and does not necessarily represent the official views of the agencies funding the research.
\end{acks}

\bibliographystyle{ACM-Reference-Format}
\bibliography{bibfile}

\input{sections/appendix}

\end{document}

%% file: sections/abstract.tex
\begin{abstract}
Explainable Artificial Intelligence (XAI) aims to uncover the decision-making processes of AI models. However, the data used for such explanations can pose security and privacy risks. Existing literature identifies attacks on machine learning models, including membership inference, model inversion, and model extraction attacks. These attacks target either the model or the training data, depending on the settings and parties involved.

XAI tools can increase the vulnerability of model extraction attacks, which is a concern when model owners prefer black-box access, thereby keeping model parameters and architecture private. To exploit this risk, we propose AUTOLYCUS, a novel retraining (learning) based model extraction attack framework against interpretable models under black-box settings. As XAI tools, we exploit Local Interpretable Model-Agnostic Explanations (LIME) \bl{and Shapley values (SHAP)} to infer decision boundaries and create surrogate models that replicate the functionality of the target model. LIME \bl{and SHAP are} mainly chosen for \bl{their} realistic yet information-rich explanations, coupled with \bl{their} extensive adoption, simplicity, and usability.

We evaluate AUTOLYCUS on six machine learning datasets, measuring the accuracy and similarity of the surrogate model to the target model. The results show that AUTOLYCUS is highly effective, requiring significantly fewer queries compared to state-of-the-art attacks, while maintaining comparable accuracy and similarity. We validate its performance and transferability on multiple interpretable ML models, including decision trees, logistic regression, naive bayes, and k-nearest neighbor. Additionally, we show the resilience of AUTOLYCUS against proposed countermeasures.
\end{abstract}

%% file: sections/intro.tex
\section{Introduction}
\label{sec:intro}

Machine Learning (ML) has become a crucial tool for many businesses, enabling them to analyze data and make predictions with greater accuracy and efficiency. 
To make ML accessible to a wider range of businesses, cloud service providers offer \bl{cost-effective} Machine Learning as a Service (MLaaS) platforms, providing customers with pretrained models or tools to deploy their own models. 

As the adoption of MLaaS platforms grew, there has been a corresponding increase in the demand for tools that facilitate explainable Artificial Intelligence (XAI)~\cite{BARREDOARRIETA202082}. 
These tools are crucial in providing users with transparency and a comprehensive understanding of how decisions are made by ML models~\cite{ibm}. 
The lack of interpretability and transparency in models that inherently operate as black boxes, such as neural networks, or interpretable models operating in black-box settings--which we specifically focus on in this paper--(i.e., where users have limited or no access to the internal workings of a model) can foster distrust and hinder their adoption.

\bl{Local Interpretable Model-agnostic Explanations (LIME)~\cite{10.1145/2939672.2939778} and SHapley Additive exPlanations (SHAP) are two popular XAI techniques utilized by major MLaaS platforms such as Azure Machine Learning (Responsible AI)~\cite{azureExp}, Watson Machine Learning (OpenScale)~\cite{ibmExp}, Amazon SageMaker (Clarify)~\cite{awsClarify}, and Google Cloud (Vertex Explainable AI)~\cite{googleCloud}.}
LIME generates local explanations for any ML model prediction by creating a new model around the sample of interest. 
Whereas, SHAP computes global feature importance values using Shapley values~\cite{shap} from cooperative game theory~\cite{owengame} and a local approximation of the model to compute feature importance values for the entire dataset.
 
The monetization of MLaaS and increasing reliance on the models provided therein have opened up new risks for businesses. 
One such risk is the threat of inference attacks including membership inference~\cite{shokri2017membership}, model inversion~\cite{10.1145/2810103.2813677}, and model extraction~\cite{10.5555/3241094.3241142}. 
In particular, threat of model extraction (or stealing) attacks proposed by Tramer et al.~\cite{10.5555/3241094.3241142} on MLaaS platforms can result in theft of proprietary models, valuable intellectual property, and the loss of monetary gains. 
In such attacks, an adversary attempts to extract the target ML model, either in whole or in part, by training a surrogate model using a set of queries to the target model. 
In addition, an attacker can execute a model extraction attack more easily using the information provided by XAI tools. 
\bl{For example}, Milli et al.~\cite{10.1145/3287560.3287562} introduced the first model extraction attack that exploits explanations \bl{to solve the hyperplanes in} neural network image classifiers with \bl{the help of} gradient based explanations. 
Chandrasekaran et al.~\cite{chandra} demonstrated that the model extraction attacks can be formulated as active learning. 
In addition to privacy concerns, Rudin~\cite{rudin2019stop} proposed that the use of interpretable models should be increased instead of explaining black-box \bl{(non-interpretable)} models for high-stake decisions. 
This is to ensure accountability and facilitating human understanding and trust in AI systems.

\bl{Although the exploitation of AI explanations for adversarial machine learning is a recent area of investigation, there have been several studies \cite{10.1145/3287560.3287562,Yan,Miura,Truong,kariyappa} conducting model extraction attacks with XAI guidance on neural networks and image classification tasks. These studies predominantly rely on image classifier-specific explanation tools (e.g., class activation maps (CAMs)~\cite{zhou2016learning, selvaraju2017grad}), which provide richer, gradient-based information compared to model-agnostic explanations. While insightful, the exclusive focus on neural networks and image classifiers within these studies overlooks the potential insights and advancements that could be gained from exploring alternative model architectures and diverse data settings. Reliance solely on explanation methods suitable for neural networks may be incompatible with certain model types or inadvertently disclose excessive model information for the less complex interpretable models, effectively transforming the attacks into semi-white-box scenarios. In contrast, our objective is to conduct a model-agnostic attack with minimal interference across multiple interpretable models. Besides, the studies targeting interpretable models, which are still highly used in ordinary statistical modeling and data analysis due to their simplicity, are limited. Thus, exploring the vulnerabilities that AI explanations present to such models, especially under black-box settings, is an uncharted territory.}

In this work, we propose a novel, model-agnostic, retraining (learning) \bl{and perturbation} based model extraction attack framework called AUTOLYCUS that exploits XAI to target interpretable models in black-box access settings. As a prior-guided adversarial-example crafting method, AUTOLYCUS leverages model explanations on local decision boundaries \bl{and feature importances. SHAP offers feature importances while LIME offers both feature importances and decision boundaries. This} can be exploited to generate membership queries.

AUTOLYCUS is influenced by the line-search retraining and adaptive retraining methods, designated as the main strategies in label-only scenarios, as proposed by Tramer et al.~\cite{10.5555/3241094.3241142}.
\bl{Line-search retraining} sends adaptive membership queries to the vicinity of the decision boundaries of a model based on previous queries, contrary to the uniform queries of other label-only attacks~\cite{choquette,10.1145/3052973.3053009}. AUTOLYCUS differs from \bl{the aforementioned attacks} in terms of how it utilizes the explanations. \bl{AUTOLYCUS does not directly seek decision boundaries. Instead, it utilizes line-search within the vicinity of explanations to generate samples that accurately reflect the decision boundaries during the retraining. While AUTOLYCUS can leverage external resources such as additional datasets or linkage data as prior information for the adversary, we limit access to black-box and constrain the information disclosed by explanations. While approaches like Equation Solving and Path-finding~\cite{10.5555/3241094.3241142} suffer from auxiliary white-box model information and realism, our approach imposes lighter assumptions on the adversary compared to similar works~\cite{10.1145/3287560.3287562,10.1145/3052973.3053009,10.5555/3241094.3241142} retaining practicality and realism.}

We demonstrate the effectiveness of AUTOLYCUS on different ML datasets in terms of the accuracy of the constructed surrogate model and the similarity of the surrogate model to the target one. We show that AUTOLYCUS provides high accuracy and similarity for interpretable models like decision trees, logistic regression, naive bayes, and k-nearest neighbor. Additionally, we demonstrate that AUTOLYCUS outperforms the SOTA methods and remains effective even in the presence of countermeasures.

\noindent\textbf{Contributions} Main contributions of AUTOLYCUS are as follows. 
\begin{itemize}
\item AUTOLYCUS \bl{can be initiated with} minimal prerequisites (e.g., minimal hyperparameter knowledge and auxiliary data), making it an easily deployable and efficient \bl{attack}. 
\item \bl{AUTOLYCUS can produce high-fidelity and high-accuracy reconstructions of the target model with fewer queries compared to the exact reconstruction attacks such as Equation-Solving~\cite{10.5555/3241094.3241142}, Path-Finding~\cite{10.5555/3241094.3241142}, and Lowd-Meek~\cite{lowd}.}
\item AUTOLYCUS is a retraining based attack. It can create partially extracted surrogate models with comparable accuracy and similarity even when it is under hard query limitations.
\item To the best of our knowledge, AUTOLYCUS is the first model extraction attack that exploits AI explanations for targeting interpretable models \bl{with black-box access}.
\item To the best of our knowledge, AUTOLYCUS is the first model extraction attack that exploits model agnostic AI explanation tools.
\end{itemize}

The rest of the paper is organized as follows. In the next section, we go over the related work. In Section~\ref{sec:xai}, we provide background on XAI. In Section~\ref{sec:systhreatmodel}, we describe the considered system and threat models. In Section~\ref{sec:methodology}, we describe AUTOLYCUS \bl{system model} in detail. In Section~\ref{sec:eval}, we provide the evaluation results. In Section~\ref{sec:countermeasures}, we show the performance of AUTOLYCUS in the existence of potential countermeasures. In Section~\ref{sec:discussion}, we discuss potential extensions and limitations. Finally, in Section~\ref{sec:conclusion}, we conclude the paper.

%% file: sections/related_work.tex
\section{Related Work}
\label{sec:relatedwork}

We review two primary lines of related research: (i) privacy attacks in machine learning and (ii) privacy risks due to model explanations. 

\noindent\textbf{Privacy attacks in ML.} Research on machine learning privacy has identified several attacks, such as membership inference~\cite{shokri2017membership,salem2018ml,10.1145/3460120.3484575,choquette} to determine whether a given user is part of the training dataset; attribute inference~\cite{ganju2018property,song2019overlearning} to infer additional private attributes of a user based on the observed ones; model inversion~\cite{10.1145/2810103.2813677,carlini2019secret,zhang2020secret} to reconstruct data from a training dataset; and model extraction~\cite{10.5555/3241094.3241142} to infer the model parameters/hyperparameters and/or the architecture of a model. 
In this work, we focus on model extraction (or stealing) attacks where an attacker tries to extract sensitive information or intellectual property from a trained model. 
The attacker typically has access to the model's input-output behavior and uses this information to build a replica of the model (functionality extraction)~\cite{7943475, orekondy2019knockoff} or extract other valuable information from it (attribute extraction)~\cite{10.5555/3241094.3241142, wang2018stealing, duddu2018stealing, 8806737}. 
Tramer et al.~\cite{10.5555/3241094.3241142} propose the first model stealing attack considering both white-box and black-box access to ML models. 
Wang et al.~\cite{wang2018stealing} propose a model extraction attack that targets the hyperparameters of ML models.
Papernot et al.~\cite{10.1145/3052973.3053009} propose a model extraction attack against black-box DNN classifiers. 
Such attacks are effective against a variety of machine learning models, including decision trees, deep neural networks, and support vector machines~\cite{10.1145/3287560.3287562,10.5555/3241094.3241142,251526, wang2018stealing}.

\noindent\textbf{Privacy risk due to model explanations.} While XAI techniques are useful for the interpretation of the prediction provided by the machine learning models, they can also be leveraged to enhance privacy attacks. 
Shokri et al.~\cite{shokri2021privacy} study the privacy risk due to membership inference attacks based on feature-based model explanations and show how various explanation methods leak information about the training data. 
Zhao et al.~\cite{zhao2021exploiting} analyze how explanation methods enhance model inversion attacks. Their work focuses on image datasets like MNIST~\cite{lecun-mnisthandwrittendigit-2010} and CIFAR-10~\cite{cifar}, and shows that the accuracy of the model inversion attacks increases significantly when explanations are used.
Kariyappa et al.~\cite{kariyappa}, Milli et al,~\cite{10.1145/3287560.3287562}, Miura et al.~\cite{Miura}, Truong et al.~\cite{Truong}, and Yan et al.~\cite{Yan} propose various model extraction attacks by exploiting gradient based AI explanations (e.g., Saliency Maps~\cite{simonyan2013deep}, Class Activation Maps~\cite{zhou2016learning}) on black-box image classifiers.
A\"ivodji et al.~\cite{aivodji} utilize counterfactual explanations for model extraction attacks.
In this paper, in contrast to Shokri and Zhao's works, we focus on model extraction attacks \bl{exploiting explanations}.

%% file: sections/preliminaries.tex
\section{Background on Explainable AI}
\label{sec:xai}
Explainable AI (XAI) is an approach for developing artificial intelligence systems that are transparent, interpretable, and can provide clear explanations for their decision-making processes~\cite{holzinger2022explainable, dovsilovic2018explainable}. The goal of XAI is to ensure that AI systems are trustworthy, fair, and accountable, and that they can be used to inform human decision-making in a reliable and understandable manner. Traditional AI algorithms, like deep learning neural networks, can be very effective at making decisions based on large amounts of data, but they are often viewed as ``black boxes'' because their internal workings are difficult to understand. This can lead to concerns about bias, discrimination, or errors in decision-making~\cite{rudin2019stop}. XAI aims to address these concerns by integrating techniques for transparency and interpretability into the design of AI systems, including those that are already deemed interpretable. This may involve using simpler models, providing visualizations or natural language explanations, or detecting anomalous behavior.

In the literature, various XAI techniques have been developed for improving the interpretability and transparency of machine learning models, particularly for neural network models in computer vision. Examples of such techniques include Layer-wise Relevance Propagation (LRP)~\cite{bach2015pixel}, Deep Taylor Decomposition (DTD)~\cite{kindermans2017learning}, Pattern Difference Analysis (PDA)~\cite{zintgraf2017visualizing}, Testing with Concept Activation Vectors (TCAV)~\cite{kim2018interpretability}, and Explainable Graph Neural Networks (XGNN)~\cite{yuan2020xgnn}. For interpretable models such as decision trees, XAI techniques such as Anchors~\cite{ribeiro2018anchors}, SHAP~\cite{shap}, and LIME~\cite{10.1145/2939672.2939778} can be utilized to provide explanations.

For further information about the techniques utilized in major MLaaS platforms and why LIME and SHAP are primarily focused in this paper, please refer to the Appendix.

\noindent{\textbf{Local Interpretable Model-Agnostic Explanations (LIME)}}
\label{sec:lime}
LIME \cite{10.1145/2939672.2939778} is a widely used XAI tool by model owners to provide an understanding of the rationale behind machine learning models predictions. This is accomplished by generating local approximations to the models around a specific instance of interest or a sample. For instance, To provide explanations for a sample $X$ on model $M$, LIME creates a dataset of different $\hat{X}$s which are the perturbed versions of $X$. This perturbation is executed differently depending on the data type, such as through pixel masking in image data or word replacement in text data. Then, each sample in the perturbed dataset is classified by $M$. Following this, LIME fits a local linear model around $X$, using $\hat{X}$s and their classifications by $M$. The resulting linear model has its own local decision boundaries to explain both linear and non-linear boundaries of $M$, allowing samples to be created and exploited within its vicinity. The feature weights in the local linear model denote the relative importance of each feature in $M$'s prediction, presented through means such as a heat-map or feature importance scores. 
It is important to note that the decision boundaries of the local linear model generated by LIME might not exactly match $M$'s exact decision boundaries due to non-linearity. But, for interpretable models, these local explanations are more accurate and consistent compared to the explanations provided to black-box models like neural networks.

\noindent{\textbf{Shapley Values (SHAP)}}\label{sec:shap}
\bl{SHAP utilizes Shapley values~\cite{shapley} of a sample to determine the marginal contribution of each feature. 
These values are then presented as explanations of that particular sample on a given prediction. 
In predictions, information released by SHAP can vary drastically. 
For instance, it can either only release the corresponding Shapley values per class - which is the most realistic scenario considering black-box access - or it can provide a plot that displays every sample in the training dataset along with the particular sample. 
Needless to say, the latter leaks a lot of information about the model. 
In AUTOLYCUS, we assume the former. 
To integrate SHAP into AUTOLYCUS, we either need to transform Shapley values into decision boundaries (like LIME) or only computing the feature importance based on them.
Given the undirected and scalar structure of SHAP explanations, decision boundaries can not be derived. Hence, we utilize feature importances of SHAP.}

\begin{figure}[ht]
\centering
\includegraphics[width=\columnwidth]{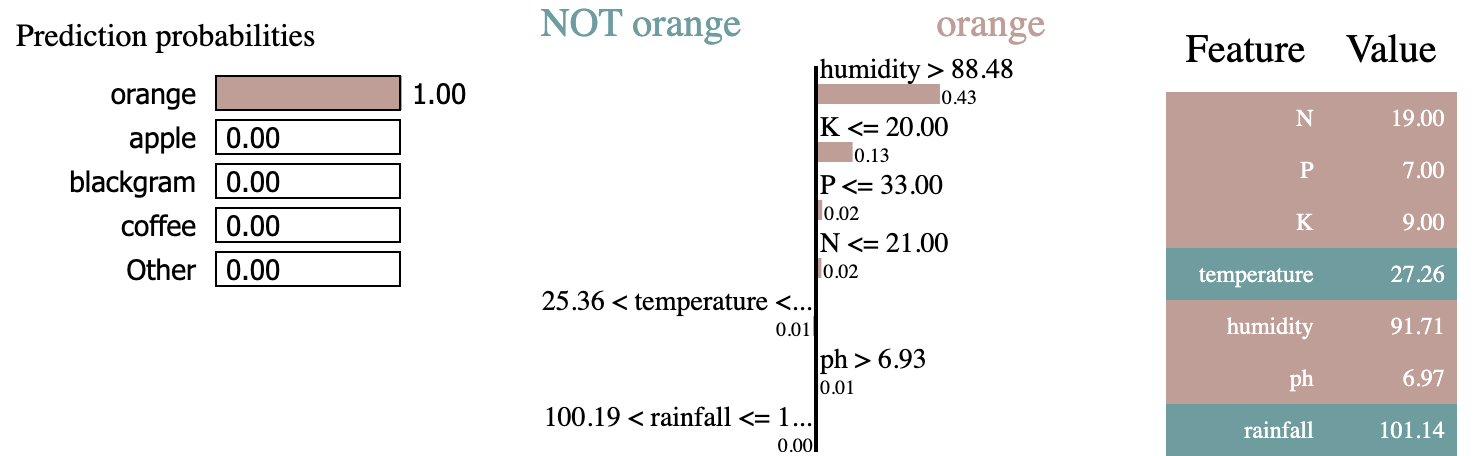}
\caption{LIME explanation example from the Crop dataset}
\Description[This is an example LIME explanation for a sample from Crop dataset.]{This is an example LIME explanation for a sample from Crop dataset. At the left hand side, ML model's prediction probabilities of the sample is given. At the middle, influential features contributing to that particular probabilities are sorted from top to bottom in descending order. At the right hand side, each feature and its value is displayed with color coding. Each color code represents if the feature positively or negatively influenced the model's decision.}
\label{fig:lime_figure}
\end{figure}

\begin{figure}[ht]
\centering
\includegraphics[width=\columnwidth]{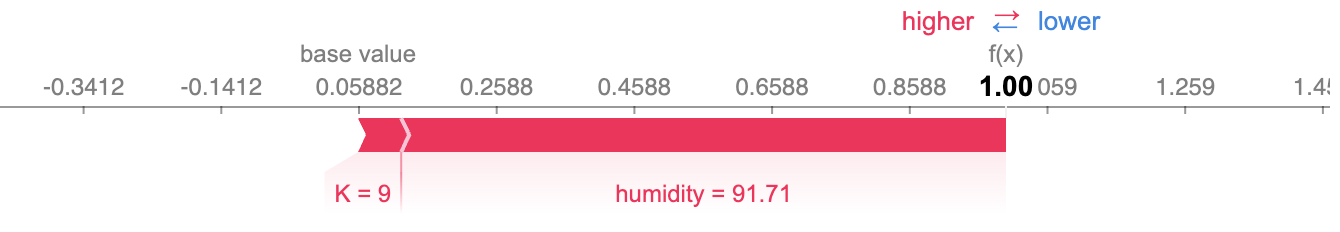}
\caption{SHAP explanation example from the Crop dataset}
\Description[This is an example SHAP explanation for a sample from Crop dataset.]{This is an example SHAP explanation for a sample from Crop dataset. The figure demonstrates which features direct the model from its regular expected value into the model's particular prediction. In this sample, humidity and potassium feature values are the strongest contributor for model's orange prediction.}
\label{fig:shap_figure}
\end{figure}

We present illustrative examples of LIME \bl{and SHAP} applied to a decision tree model trained on the Crop dataset~\cite{crop}. Figure~\ref{fig:lime_figure} displays the LIME explanation of a membership query which is classified as orange by the target model. On the left-hand side, the prediction probabilities for each class are presented. 
In the middle section of the figure, the local factors contributing to the model's decision are listed. These factors are ranked in decreasing order of their explanatory strength. Each local factor is composed of contributing features and their corresponding decision boundary intervals. On the right-hand side, the features and their values are listed, color coded to indicate whether they strengthen or weaken the model's prediction.
Figure~\ref{fig:shap_figure} \bl{displays the SHAP explanation of the same sample with feature importances.}
\bl{These examples are} intended to demonstrate the functionality of LIME \bl{and SHAP} and the type of information they provide for an interpretable model. Further details about the Crop dataset and the experimental setup are presented in Section~\ref{sec:datasets} and subsequent sections.

%% file: sections/methodology.tex
\section{System and Threat Models}
\label{sec:systhreatmodel}

\begin{figure*}[!ht]
\centering
\includegraphics[width=\textwidth, height=0.70\columnwidth]{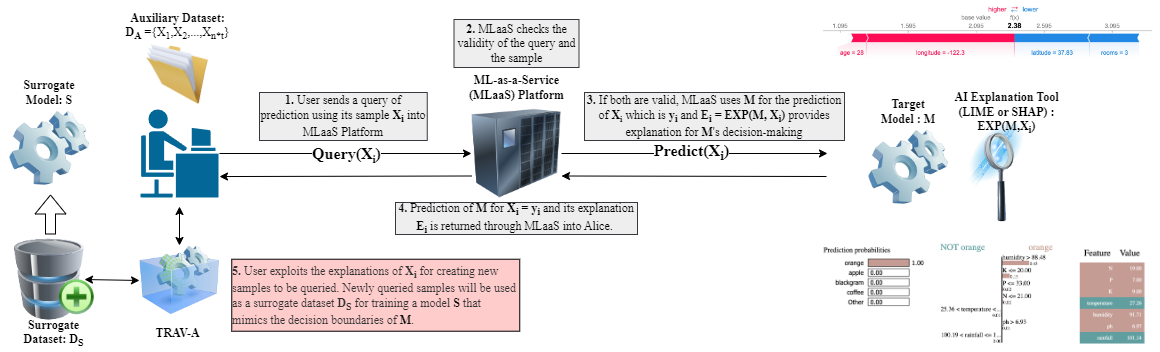}
\caption{AUTOLYCUS system diagram consisting of the following steps: (1) a user sends a query to the MLaaS platform, (2) the MLaaS platform verifies the validity of the query such that no empty or incomplete queries are sent, (3) the ML model $M$ predicts the class of the queried sample $y_i$ and \bl{the explainer} computes its explanation $E_i$, (4) the MLaaS platform returns the results to the user, and (5) in case of an adversarial user, they exploit explanations via TRAV-A algorithm (as described in Section~\ref{sec:generation}) to extract the decision boundaries of the target model $M$.} 
\Description[This is the system diagram of AUTOLYCUS.]{AUTOLYCUS system diagram consisting of the following steps: (1) a user sends a query to the MLaaS platform, (2) the MLaaS platform verifies the validity of the query such that no empty or incomplete queries are sent, (3) the ML model $M$ predicts the class of the queried sample $y_i$ and \bl{the explainer} computes its explanation $E_i$, (4) the MLaaS platform returns the results to the user, and (5) in case of an adversarial user, they exploit explanations via TRAV-A algorithm (as described in Section~\ref{sec:generation}) to extract the decision boundaries of the target model $M$.}
\label{fig:system_model_figure}
\end{figure*}

Consider a target model $M$ that is trained on a dataset $D_{M}$ and is deployed for use via an API (e.g., in an ML-as-a-Service (MLaaS) platform). Users can send queries to the target model $M$ and receive as an output, the predicted label, and the corresponding explanation generated by \bl{the explainable AI (XAI) tools} (as described in Section~\ref{sec:lime}). 
For instance, a user sends a query for a given sample $X_i$ to the target model $M$ and receives an output including the predicted class $y_i$, and its corresponding explanations $E_i$ provided by \bl{XAI} (as shown in Figure~\ref{fig:system_model_figure}).
A model extraction attack occurs when an adversary attempts to learn a surrogate model $S$ that closely approximates $M$ by exploiting both the output of the predicted labels and the explanations returned by \bl{XAI tools}. The goal of the attacker is to \bl{create a surrogate dataset that inherently reflects the decision boundaries, and hence the functionality of the target model} $M$.

We assume that the attacker possesses black-box access to the target model $M$, including knowledge of \bl{the type -not architecture- of the machine learning model (i.e., decision tree, logistic regression)}.
Even unknown, the model type can be inferred with reasonable accuracy by an individual possessing expertise in machine learning transferability~\cite{papernot2016transferability} during retraining.
Furthermore, the attacker is informed of the possible values that $y_i$ may take, i.e., the class names, and is thus aware of the total number of classes $t$.
The attacker has access to an auxiliary dataset $D_A$, comprising one or multiple samples to initialize the traversal algorithm. This is to prevent sending blind queries resulting in overhead to the query budget. The composition of $D_A$ may range from a rudimentary collection of samples to comprehensive datasets. 
Note that in the absence of samples, a random query of default sample values can still be utilized to initiate the process. 
\bl{This can be utilized to simulate a scenario where the attacker lacks auxiliary information.} However, the exploration capability with such limited samples \bl{can be} restrictive, and a higher query budget is essential for an efficient model extraction attack compared to scenarios where a more extensive $D_A$ is available.
Lastly, the attacker knows the basic characteristics of the features of the dataset that is used to train model $M$, such as their type, domain, and lower and upper bounds (i.e., 0 $\leq$ pH $\leq$ 14). 

We direct the reader's attention to Table~\ref{tbl:symbols} for the symbols and notations that appear frequently in this paper.

\begin{table}[!ht]
\caption{Frequently used symbols and notations.}
\centering
\renewcommand{\arraystretch}{1.3}
\resizebox{\columnwidth}{!}{
\begin{tabular}{|c|l|}
\hline
$M$             &   Target Model                        \\ \hline
$S$             &   Surrogate Model                     \\ \hline
$D_{M}$         &   Training dataset of target model    \\ \hline
$D_{S}$         &   Training dataset of surrogate model \\ \hline
$D_{A}$         &   Attacker's auxiliary dataset        \\ \hline
$D_{E}$         &   Exploration dataset                 \\ \hline
$Q$             &   Query budget (number of queries)    \\ \hline
$X_i$           &   Sample $i$                          \\ \hline
$\hat{X_i}$     &   Altered sample $i$                  \\ \hline
$x_i^j$         &   Sample $i$'s $j^{th}$ feature       \\ \hline
$\hat{x}_i^j$   &   Altered feature $j$ of sample $i$   \\ \hline
$y_i$           &   Predicted class for sample $i$      \\ \hline
$E_i$           &   Explanation for sample $i$          \\ \hline
$Exp_M$         &   Explainer function of model $M$     \\ \hline
$k$             &   Number of top features allowed to be explored \\ \hline
$m$             &   Number of total features            \\ \hline
$n$             &   Number of samples per class         \\ \hline
$t$             &   Number of total classes             \\ \hline
$l,u$           &   Lower and upper bounds for the number of samples generated per class \\ \hline
$db_i^j$        &   Decision boundaries of the sample $i$'s $j^{th}$ feature \\ \hline
$\delta$        &   Alteration coefficient(s) \\ \hline
TRAV-A          &   The traversal algorithm \\ \hline
\end{tabular}
}
\label{tbl:symbols}
\end{table}

\section{Proposed Work}
\label{sec:methodology}

In AUTOLYCUS, the attacker first needs to create a surrogate dataset $D_S$ to learn a surrogate model $S$ that closely approximates the target model $M$. We assume that the attacker has access to $n$ samples per class in the auxiliary dataset $D_A ={X_1, X_2, \dots, X_{t*n}}$ which may or may not have samples from the original dataset $D_{M}$. Here, $X_i=\{x_i^1, x_i^2, \dots, x_i^m$\}, where $x_i^j$ represents the value of feature $j$ in sample $X_i$ and $m$ is the total number of features.  
We also assume that there exists a query budget $Q$ which restricts the total number of queries that an attacker can send to the target model $M$. 
The proposed model extraction attack is depicted in Figure~\ref{fig:system_model_figure}. 
In Step 1, the attacker sends a query using one of its local samples (from the auxiliary dataset) to generate additional samples. In Step 2, the validity of the query is confirmed. In Step 3, the ML model $M$ predicts the class of the queried sample, and LIME \bl{or SHAP} computes its explanation. In Step 4, the MLaaS platform returns the \bl{model's class} prediction along with its explanation to the user. In the last step (Step 5), the attacker parses and uses the model explanations to generate altered samples that differ from their predecessors by a single \bl{or $k$ features}. \bl{SHAP returns the Shapley values of each feature for the class predicted by $M$ and they are parsed as feature importances.} 
This process is informed and guided by explanations as opposed to sending random queries. By increasing the number of queries sent, more samples labeled with the target model's predictions are retrieved.  
\bl{In any retraining attack, the samples utilized as the training dataset $D_S$ for surrogate models $S$ are intended to mimic the behavior of the target model.}
In the following, we describe the details of AUTOLYCUS.

\subsection{Generating Candidate Samples}
\label{sec:generation}

The goal of the attacker is to send informative queries to the target model $M$ instead of blind queries. 
For that, depending on the XAI tool that is used by the target model, the attacker explores the decision boundaries returned by LIME explanations \bl{or feature importances returned by SHAP explanations}. 
Assume the attacker sends a query for a given sample $X_i$ (e.g., one of the samples in $D_A$) to the target model $M$. 
As discussed, the target model $M$ sends the predicted class $y_i$ and the corresponding explanation $E_i$. 
As discussed in Section~\ref{sec:lime}, the explanation returned by LIME consists of the prediction \bl{probabilities (due to black-box access scenario, we assume only the top class)} $y_i$ and the decision boundaries $db_i^j$ ordered by feature importances. \bl{Whereas, SHAP only returns the feature importances.}
To generate new and informative candidate samples, denoted as $D_C$ in Algorithm~\ref{alg:trava}, the attacker considers only the top $k$ features of the sample $X_i$ \bl{through its feature importances}.
\bl{There are two strategies to follow here. In the first exploration procedure, the features are altered one by one resulting in a breadth-first tree search} with the width and depth of the tree depending on the chosen value of $k$. \bl{In the second exploration procedure, the top $k$ features are altered all at once. 
This results in a more diverse set of samples explored. This is an optimal strategy to follow if the size of $D_A$ is small and the distribution of classes is close to uniform distribution. However, if the size of $D_A$ is large and if there are rare classes present, the first procedure may provide better exploration results. In the first exploration procedure,} a higher $k$ value will result in a wider tree with low depth, implying that there are more features, and thus more variants of the same sample that need to be explored. 
This may lead to the exploration of numerous uninformative samples and ultimately exhaust the query budget. 
Conversely, a lower $k$ value will yield a slender tree with high depth, indicating that more samples are examined with only the top features being prioritized. 
This results in the inadequate exploration of informative samples, potentially losing crucial information about the structure of the tree.
The choice between a high or low value of $k$ \bl{and altering them one-by-one or all-at-once} as a viable strategy depends on the distribution of feature importances (if they are known or inferred). 

For each of the top $k$ features, the attacker generates candidate samples by analyzing the decision boundaries returned by LIME \bl{or the feature importances returned by SHAP}. 
The attacker computes new values of feature(s) $j$, denoted as $\hat{x_i}^j$ by altering it into the decision boundary $db_i^j$ in LIME \bl{or to the next available value in SHAP} with the coefficient(s) denoted as $\delta_j$. 
Formally, $\hat{x_i}^j$ is computed as: 
$\hat{x_i}^j=db_i^j \pm \delta_j$ (in LIME) or $\hat{x_i}^j = {x_i}^j \pm \delta_j$ (in SHAP), where $j$ is the index of the feature that the attacker is aiming to modify and $\delta_j$ is the alteration coefficient. For LIME, $\delta$ is equal to $1$ for categorical features to reflect encoding difference and to $0.01$ (or lower) for continuous features. \bl{Since the perturbation is decided by the decision boundaries in LIME, $\delta$ has lower importance. On the other hand, it is very important in SHAP, since it determines the exploration difference between successive samples. A good rule of thumb is setting it close to the standard deviations if available or to the quarters of solution range. Depending on the intended alteration, $\delta_j$ can be manually configured to larger or lower values if necessary.} 
\bl{The resulting candidate sample is obtained as $\hat{X_i}=\{x_i^1, x_i^2, \dots, \hat{x_i}^{j_1}, \dots, x_i^m\}$ or $\hat{X_i}=\{x_i^1, \hat{x_i}^{2}, \dots,{x_i}^{m-2}, \hat{x_i}^{m-1}, x_i^m\}$.}

Figure \ref{fig:lime_figure} shows $X_i = \{19, 7, 9, 27.26, 91.71, 6.97, 101.14\}$, a toy sample from the Crop dataset~\cite{crop} and its corresponding explanation $E_i$. In this example, the top features for $k=3$ are `humidity', `K', and `P'. If we focus on `humidity' and select $\delta = 0.01$, the new generated sample will be $\hat{X_i} = \{19, 7, 9, 27.26, \textbf{88.49}, 6.97, 101.14\}$.
\bl{Figure \ref{fig:shap_figure} shows the same example for SHAP. In this example, the only influential features in the model's decision are `K' and `humidity'. When both of these features are altered by 10 ($\delta_2 = 10$ and $\delta_4 = 10$), the new generated sample(s) will be $\hat{X_i} = \{19, 7, \textbf{19}, 27.26, \textbf{91.71} \pm \textbf{10}, 6.97, 101.14\}$ ('K' $= 9 - 10 = -1$ is invalid and discarded)}

\algnewcommand\And{\textbf{ and}}
\algnewcommand\Or{\textbf{ or}}
\begin{algorithm}[!ht]
    \caption{Traversal algorithm used for generating samples that will train the surrogate model}
    \label{alg:trava}
    \begin{algorithmic}[1]
    \Function{\textbf{TRAV-A}}{$D_A$, $M$, $Exp_M$, $t$, $n$, $l$, $u$, $Q$, $\delta$}
    \State $D_E$, $D_S$, $y_S$ $\gets D_A, \left \{ \right \}, \left \{ \right \}$
    \State visits $\gets \{0,0,...0\}$ \Comment{(length $t$)}
    \State q $\gets$ 0
    \While{$D_E$ $\neq \emptyset$ \textbf{and} ($Q \geq$ q \textbf{or} ANY(visits)) $\leq l$}
    \State q++
    \State $X_i$ $\gets$ $D_E$.\texttt{POP}()
    \State $y_i$ $\gets$ $M$.\texttt{PREDICT}($X_i$)
    \If{visits[$y_i$] $< u$}
    \State visits[$y_i$]++
    \State $D_S$.\texttt{ADD}($X_i$)
    \State $y_S$.\texttt{ADD}($y_i$)
    \State $E_i$ $\gets$ \textbf{PARSE\_EXP}($X_i$, $M$, ${Exp_M}$, $t$, $n$)
    \State $D_C$ $\gets$ \textbf{GENERATE\_SAMPLES}($X_i$, $E_i$, $n$, $\delta$)
    \For{i \textbf{in} $D_C$}
    \If{i $\notin$ $D_E$, $D_S$}
    \State $D_E$.\texttt{PUSH}(i)
    \EndIf
    \EndFor
    \EndIf
    \EndWhile
    \State \textbf{return} $D_S$, $y_S$
    \EndFunction
\end{algorithmic}
\end{algorithm}

\subsection{Creating the Surrogate Dataset}\label{sec:surrogate_dataset}

In order to have a balanced dataset, the attacker aims to create a surrogate dataset $D_S$ that contains similar and proportional number of samples per class to the target model's training dataset $D_M$.
Thus, it needs to generate a minimum number of samples lower-bounded by a value denoted as $l$ per class and $l\times t$ samples in total. This is to ensure that no prediction class is under-represented in $D_S$.
In order to use the query budget $Q$ efficiently, the attacker limits the number of samples generated per class to an upper bound value denoted as $u$. 
If the bounding values of $l$ and $u$ are not enforced, the more frequently predicted classes may dominate $D_S$ during traversal, resulting in an imbalanced dataset. This imbalance may hinder the performance of resulting surrogate models $S$, leading to over-fitting in common classes and inadequate predictions for rarer classes. To address this, $l$ and $u$ values can be represented as scalar values for an equal number of samples to be discovered in each class, or as lists allowing for the exploration of different numbers of samples for each class. We recommend setting $l$ based on the frequencies of classes in $D_M$, if such information is available, and setting $u$ by incrementing $l$ with a desired margin of error. If the frequencies of classes are unknown, a good rule of thumb is setting $l$ and $u$ close to $Q/t$ such that the balance of $D_S$ is maintained for an accurate model training.     

The attacker uses the traversal algorithm \textbf{TRAV-A} to create the surrogate dataset $D_S$. This algorithm is built upon \bl{running the line-search retraining algorithm proposed by Tramer et al.~\cite{10.5555/3241094.3241142} on the model explanations}.
The detailed steps are described in Algorithm~\ref{alg:trava}. 
Recall that the attacker has access to $n$ samples per class from the auxiliary dataset $D_A$. Thus, initially, $D_S$ is limited to $D_A$. 
Let $D_E$ denote the dataset with the samples that need to be explored (initially, $D_E = D_A$). TRAV-A starts by exploring the samples in $D_E$. 
It selects the first sample $X_i$ in dataset $D_E$ and sends a query to the target model $M$. 
After receiving the predicted class $y_i$, TRAV-A checks if the maximum number of samples generated for this class has been reached. 
If that is the case, TRAV-A continues by sending a query for the next sample in $D_E$ and checking if the above condition is met. 
Otherwise, TRAV-A adds $X_i$ to the surrogate dataset $D_S$ and generates new candidate samples. 
In Section~\ref{sec:generation}, we described how to generate the candidate samples. The generated samples which have not been previously explored (i.e., is not a part of $D_E$), are added to $D_E$. 
In the next iteration, TRAV-A sends a query for the next sample in $D_E$. 
TRAV-A employs a breadth-first search strategy to traverse the candidate samples generated during the exploration of a specific sample. 
We adopt this approach to prevent the deep exploration and propagation of a single sample. 
This process continues until one of the following conditions is met: (i) there are no unexplored samples in $D_E$, (ii) the query budget $Q$ has been exhausted, or (iii) the minimum number of samples $l$ that needs to be generated per class has been reached. Note that the lower bound $l$ should be picked close to the approximate query limit per class $Q/t$ such that the budget is utilized as effectively as possible. 
It is noteworthy that the last condition (iii) is optional, but it complements the second condition (ii). This condition is included in scenarios where the query budget $Q$ is exceeded without having explored a satisfactory number of samples from each class. Consequently, if the query budget is allowed to be exceeded, condition (iii) can substitute for it and ensure that an adequate number of samples from each class have been examined. 

Using the generated surrogate dataset $D_S$, the attacker trains the surrogate model(s) $S$. Recall that the attacker knows the \bl{type} of the target model $M$ \bl{and can create multiple surrogate models under different hyperparameter settings. Then, the attacker can select the surrogate model with the most similarity without spending additional query budget. 
The attacker can calculate this similarity by refraining one-$k^{th}$ of the queried samples from the surrogate model training for $k$-fold cross-validation.} 

%% file: sections/results.tex
\section{Evaluation}
\label{sec:eval}

In this section, we describe the datasets, evaluation settings and metrics, and the obtained results. 

\subsection{Datasets}
\label{sec:datasets}
To evaluate the performance of AUTOLYCUS, we employ six common datasets, most of which are from UCI Machine Learning Repository~\cite{uci}. For details about the datasets, please refer to Table~\ref{tbl:datasets}.

\begin{table}[!ht]
\caption{Dataset properties.}
\centering
\begin{adjustbox}{max width=\columnwidth}
\begin{tabular}{|c|c|c|c|c|}
\hline
Dataset Name & \# Samples & \# Features(m) & \# Classes(t) & Data Type\\ \hline
Iris~\cite{uci} & 150 & 4 & 3 & Continuous  \\ 
Crop~\cite{crop} & 1700 & 7 & 17 & Continuous   \\ 
Breast Cancer~\cite{uci} & 569 & 30 & 2 & Continuous \\
Adult Income~\cite{uci} & 48,842 & 14 & 2 & Hybrid \\ 
Nursery~\cite{uci} & 12,960 & 8 & 3 & Categorical \\ 
Mushroom~\cite{uci} & 8,124 & 22 & 2 & Categorical \\ \hline
\end{tabular}
\end{adjustbox}
\label{tbl:datasets}
\end{table} 

\subsection{Evaluation Settings and Metrics}
\label{sec:model_param_train}

\begin{table*}[!ht]
\caption{Comparison with SOTA}
\centering
\begin{adjustbox}{max width=\textwidth}
\begin{tabular}{|c|cccc|ccccc|}
\hline
\multirow{2}{*}{Dataset} & \multicolumn{4}{c|}{SOTA Attacks} & \multicolumn{5}{c|}{AUTOLYCUS}\\
\multirow{2}{*}{} & Attack Name & Model & $1-R_{test}$ & Queries & Model & $1-R_{test}$ & Queries & $n$ & XAI Tool\\ \hline \hline
Iris & Equation Solving~\cite{10.5555/3241094.3241142} & Logistic Regression & 1 & 644 & Logistic Regression & 1 & \textbf{100} & 1 & LIME \\ \hline
Iris & Path Finding~\cite{10.5555/3241094.3241142,chandra} & Decision Tree & 1 & 246 & Decision Tree & 1 & \textbf{10} & 1 & LIME \\ \hline
Iris & IWAL~\cite{chandra} & Decision Tree & 1 & 361 & Decision Tree & 1 & \textbf{10} & 1 & LIME\\ \hline
Breast Cancer & Equation Solving & Logistic Regression & 1 & [644,1485] & Logistic Regression & 0.992 & \textbf{100} & 5 & SHAP\\ \hline
Adult Income & Equation Solving & Logistic Regression & 1 & 1485 & Logistic Regression & 0.998 & \textbf{1000} & 5 & SHAP\\ \hline
Adult Income & Path Finding & Decision Tree & 1 & 18323 & Decision Tree & 0.937 & \textbf{1000} & 5 & SHAP\\ \hline
Adult Income & IWAL & Decision Tree & 1 & 244188 & Decision Tree & 0.937 & \textbf{1000} & 5 & SHAP\\ \hline
\end{tabular}
\end{adjustbox}
\label{tbl:sota}
\end{table*}

\noindent\textbf{Evaluation settings.}
\label{sec:eval_settings}
We use scikit-learn~\cite{scikit-learn} Python package for pre-processing, training, and evaluating the models. Prior to training the models, we apply pre-processing techniques to the datasets by eliminating any missing entries and encoding categorical variables using label encoders.
Each dataset is split into three subsets: (i) the training dataset ($75\%$), (ii) the test dataset ($15\%$), and (iii) the auxiliary set ($10\%$). 
The training dataset is used for training the target model $M$, the test dataset for evaluating the performance of the model extraction attacks, and the auxiliary set for creating the auxiliary dataset $D_A$.  
We randomly generate a subset of $n$ samples per class from the auxiliary set, thereby creating a minimal dataset $D_A$. 
It is noteworthy that especially for models with a small number of classes, $D_A$ can be manually generated by the attacker \bl{with random sampling} without any prior knowledge. 
\bl{This observation closely aligns with scenarios in which the attacker lacks auxiliary data.} Then, the attacker uses $D_A$ to initiate the traversal algorithm (as described in Section~\ref{sec:surrogate_dataset}). 

We conduct experiments on the following interpretable models; decision trees, logistic regression, naive bayes, k-nearest neighbor classifiers, and random forest. We also test the applicability of AUTOLYCUS against multilayer perceptrons \bl{and discuss its limitations} in Section~\ref{sec:discussion}.
To mitigate potential bias during each model training, we train $50$ surrogate models on dataset $D_S$ using distinct random states and generating $10$ auxiliary datasets $D_A$ to initialize the traversal.
We repeat each experiment $10$ times for each auxiliary dataset $D_A$ and report the average of the results.

\noindent\textbf{Evaluation metrics.} The experiments aim to evaluate the performance of the constructed surrogate models in comparison to the target models using two distinct metrics. 
In particular, we employ the accuracy and similarity metrics to measure the efficacy of the constructed surrogate models against the test dataset. 
Accuracy represents the proportion of correct classifications relative to all predictions made by the surrogate model. 
In contrast, model similarity is the label agreement between the target model and the surrogate models against a neutral dataset, which is referred in the literature also as fidelity or $1 - R_{test}$~\cite{10.5555/3241094.3241142}. The formal definition of $R_{test}$ over a test dataset $D$ is $R_{test}(f,\hat{f})=\pmb{\sum}_{(\pmb{x},y)\in D} d(f(x),\hat{f}(x))/|D|$.   
This enables us to determine the extent to which the generated surrogate models are able to replicate the functionality of the target models.

\noindent\textbf{Comparison with other techniques.} We compare the performance of AUTOLYCUS with four other approaches.

\noindent\textit{\textbf{Baseline attack}.} The surrogate model $S$ is trained directly on the auxiliary dataset ($D_A = D_S$). In this scenario, the attacker does not send any queries ($Q=0$) to the target model $M$. 

\noindent\textit{\textbf{\bl{Steal-ML attacks}}.} Tramer et al. propose a ``path-finding attack'' to target decision tree models and an equation-solving attack to target logistic regression models for model extraction~\cite{10.5555/3241094.3241142}.  
The path-finding attack\footnote{Path-finding results are obtained from the reproduced results of Chandrasekaran et al.~\cite{chandra} and not from the original paper~\cite{10.5555/3241094.3241142}} is a deterministic, top-to-bottom attack that explores all the nodes until an exact reconstruction is achieved. 
In \cite{10.5555/3241094.3241142}, the splits in the tree path are exploited by extracting node predicates and creating alternate samples. 
Here, node predicates are conditions a sample must satisfy to end up in the same node ID. 
The downside of this attack is its strong assumptions like node IDs being readily available and incomplete queries being sent to avail exploration from the top of the tree to the bottom.
``The equation-solving attack'' is a technique employed against logistic regression models and neural networks. Its query results are converted into linear equations to be solved collectively. Depending on the setting, the extraction may require queries ranging from dozens to hundreds (can be millions for neural networks). \bl{This method uses confidence intervals as auxiliary model information, which inadvertently discloses model's internal mechanisms, thereby overlooking the context of black-box access scenarios.}

\noindent\textit{\textbf{IWAL attack}.} Chandrasekaran et al.~\cite{chandra} propose an active learning based model extraction attack IWAL to target tree structured models. IWAL is the importance weighted active learning algorithm of Beygelzimer et al.~\cite{beygelzimer}, which iteratively refines a tree in each query by minimizing the labeling error.

\begin{figure*}[!htp]  
    \begin{minipage}{.08\textwidth}
        Iris Dataset (a)
    \end{minipage}
    \hspace{0.01\textwidth}
    \begin{minipage}{.9\textwidth}
        \subfloat{\includegraphics[width=\textwidth,height=1.37in]{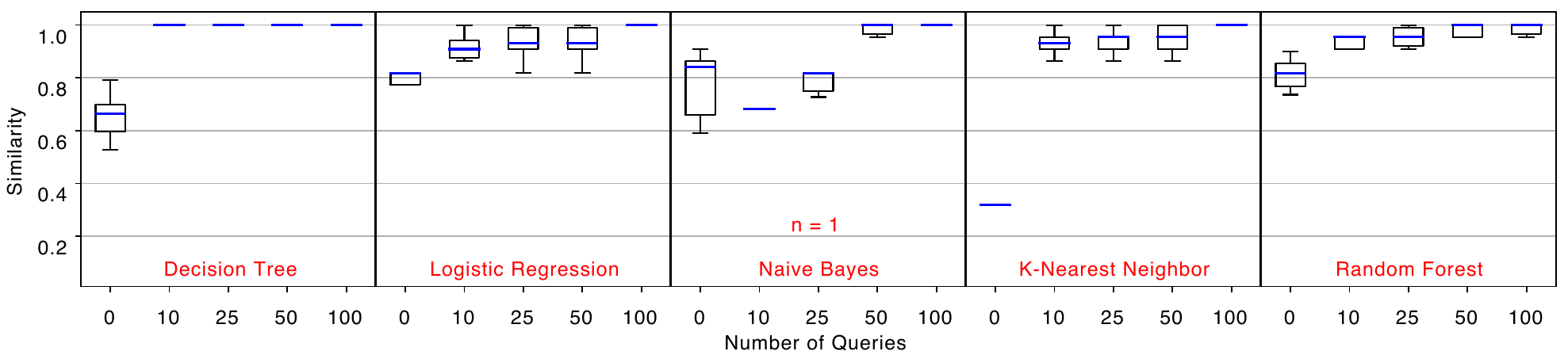}}
    \end{minipage}\\
    \begin{minipage}{.08\textwidth}
        Crop Dataset (b)
    \end{minipage}
    \hspace{0.01\textwidth}
    \begin{minipage}{.9\textwidth}
        \subfloat{\includegraphics[width=\textwidth,height=1.37in]{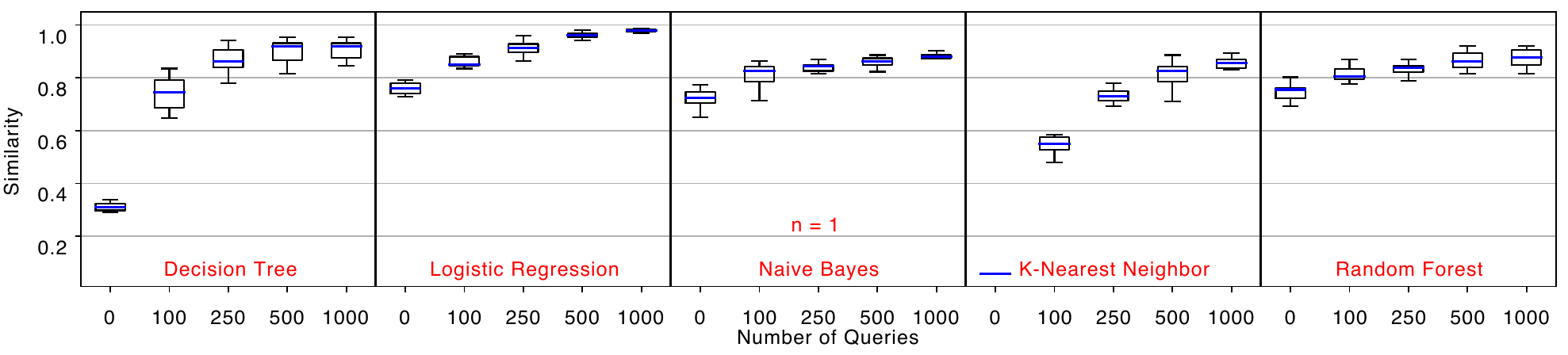}}
    \end{minipage}\\
    \begin{minipage}{.08\textwidth}
        Breast Cancer Dataset (c)
    \end{minipage}
    \hspace{0.01\textwidth}
    \begin{minipage}{.9\textwidth}
        \subfloat{\includegraphics[width=\textwidth,height=1.37in]{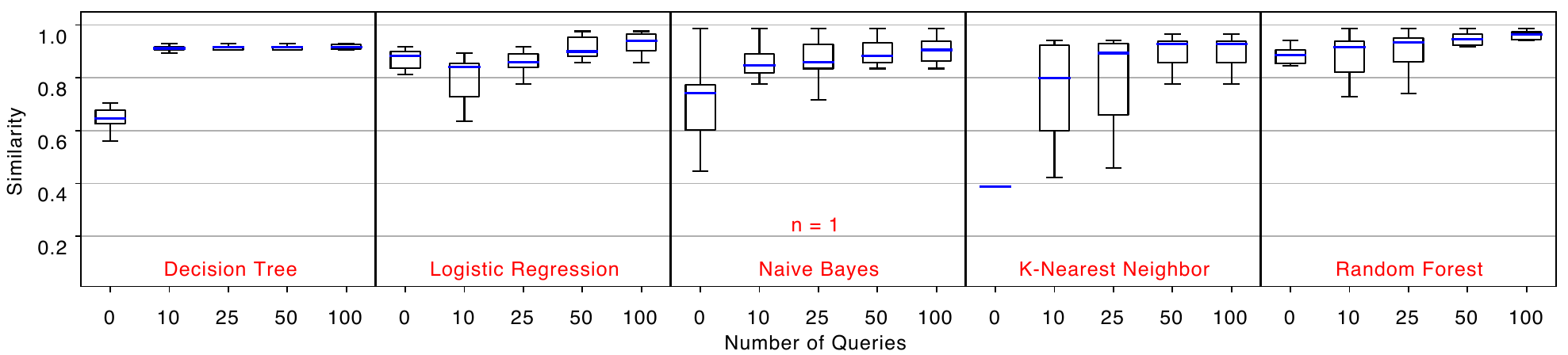}}
    \end{minipage}\\
    \begin{minipage}{.08\textwidth}
        Adult Income Dataset (d)
    \end{minipage}
    \hspace{0.01\textwidth}
    \begin{minipage}{.9\textwidth}
        \subfloat{\includegraphics[width=\textwidth,height=1.37in]{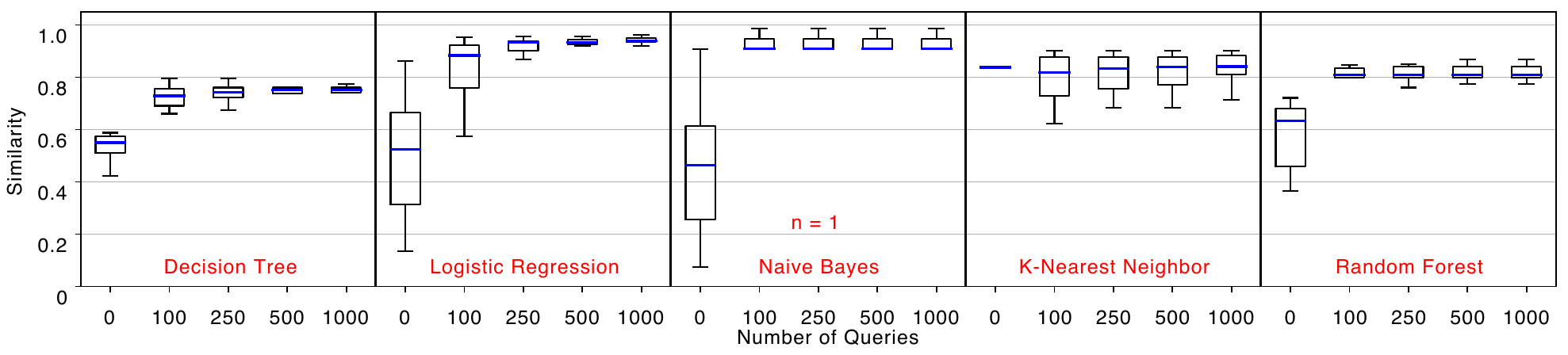}}
    \end{minipage}\\
    \begin{minipage}{.08\textwidth}
        Nursery Dataset (e)
    \end{minipage}
    \hspace{0.01\textwidth}
    \begin{minipage}{.9\textwidth}
        \subfloat{\includegraphics[width=\textwidth,height=1.37in]{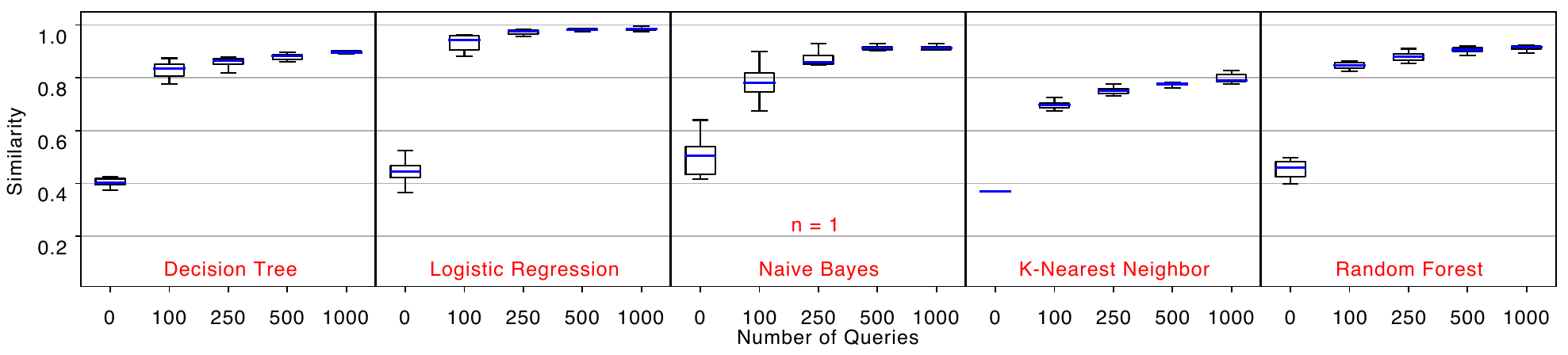}}
    \end{minipage}\\
    \begin{minipage}{.08\textwidth}
        Mushroom Dataset (f)
    \end{minipage}
    \hspace{0.01\textwidth}
    \begin{minipage}{.9\textwidth}
        \subfloat{\includegraphics[width=\textwidth,height=1.3in]{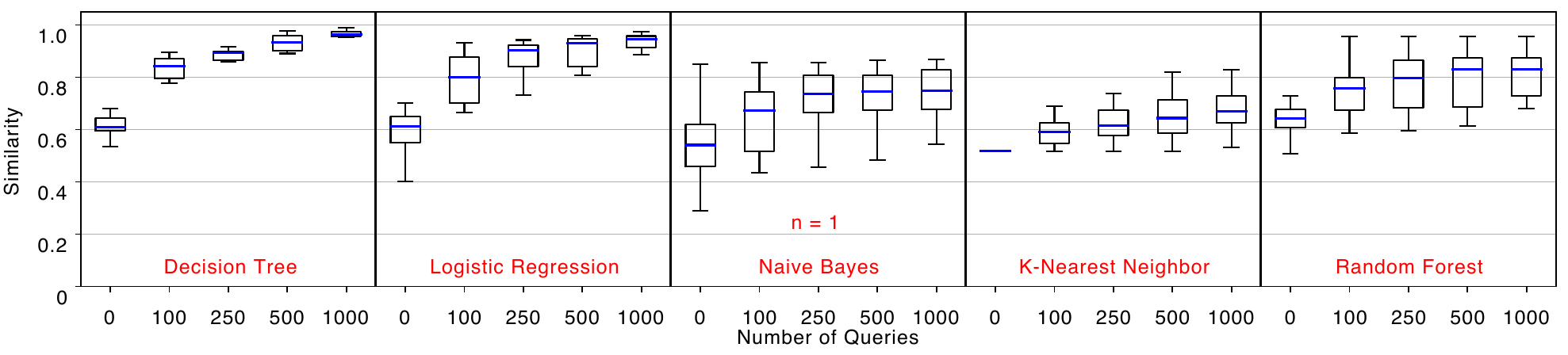}}
    \end{minipage}\\
    
    \caption{Impact of the number of queries ($Q$) on surrogate model similarity when LIME is used. $(k=3, n=1)$}
    \Description[This figure shows the experimental results of our attack when LIME is used as the explanation technique.]{This figure shows the impact of number of queries ($Q$) on surrogate model similarity when LIME is used as the explanation technique. It is a 5 row 6 column figure. Each row represents the dataset used and each column represents the ML model targeted for extraction. Datasets used are Iris (a), Crop (b), Breast Cancer (c), Adult Income (d), Nursery (e), Mushroom (f) from top to bottom. ML models targeted are Decision Tree, Logistic Regression, Naive Bayes, K-Nearest Neighbor and Random forest from left to right. Each figure has an x and y axis. X-axis represents the number of queries sent and the y axis represents the resulting surrogate model's similarity to the target model. In this figure number of features explored (k) is kept at 3 and the auxiliary dataset size per class (n) is kept at 1.}
    \label{fig:similarities1}
\end{figure*}

\begin{figure*}[!htp]  
    \begin{minipage}{.08\textwidth}
        Iris Dataset (a)
    \end{minipage}
    \hspace{0.01\textwidth}
    \begin{minipage}{.9\textwidth}
        \subfloat{\includegraphics[width=\textwidth,height=1.37in]{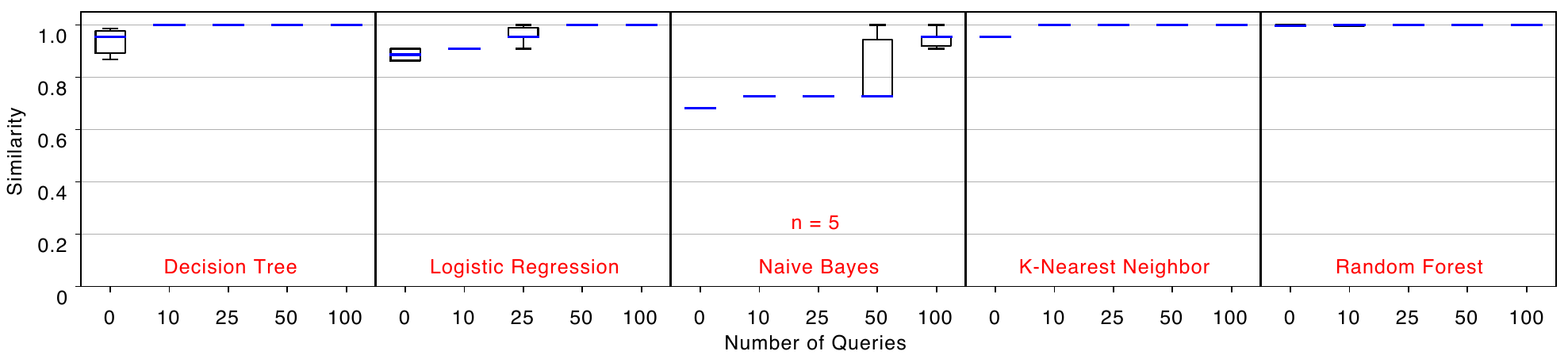}}
    \end{minipage}\\
    \begin{minipage}{.08\textwidth}
        Crop Dataset (b)
    \end{minipage}
    \hspace{0.01\textwidth}
    \begin{minipage}{.9\textwidth}
        \subfloat{\includegraphics[width=\textwidth,height=1.37in]{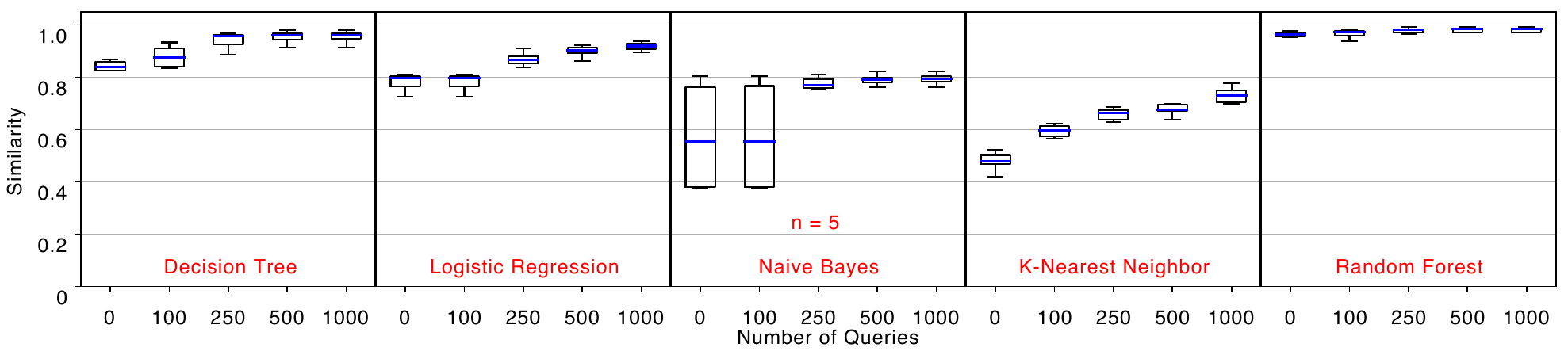}}
    \end{minipage}\\
    \begin{minipage}{.08\textwidth}
        Breast Cancer Dataset (c)
    \end{minipage}
    \hspace{0.01\textwidth}
    \begin{minipage}{.9\textwidth}
        \subfloat{\includegraphics[width=\textwidth,height=1.37in]{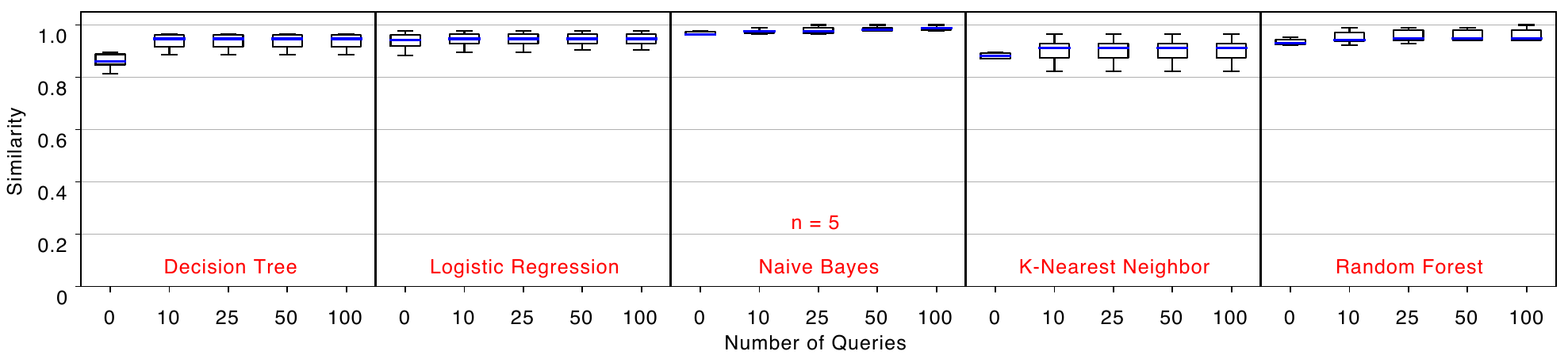}}
    \end{minipage}\\
    \begin{minipage}{.08\textwidth}
        Adult Income Dataset (d)
    \end{minipage}
    \hspace{0.01\textwidth}
    \begin{minipage}{.9\textwidth}
        \subfloat{\includegraphics[width=\textwidth,height=1.37in]{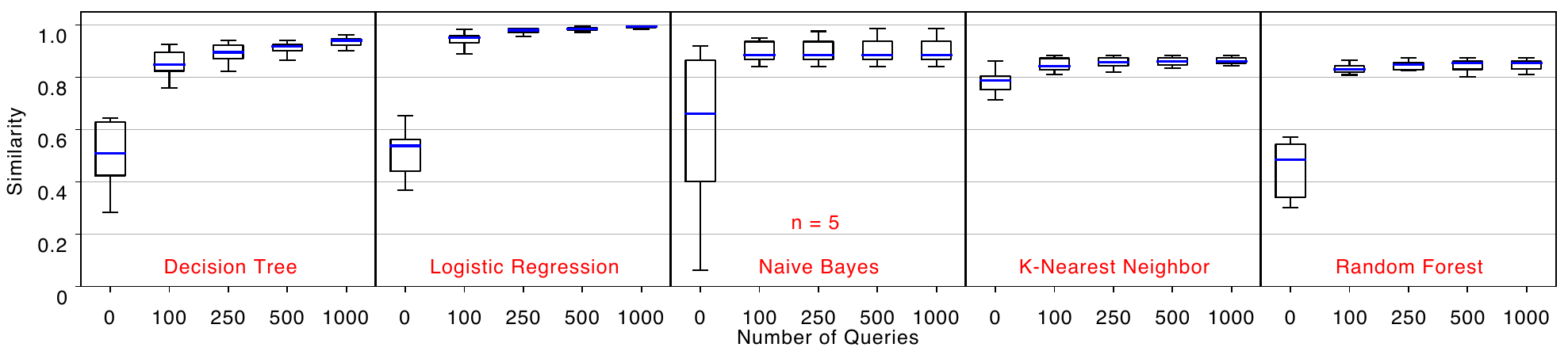}}
    \end{minipage}\\
    \begin{minipage}{.08\textwidth}
        Nursery Dataset (e)
    \end{minipage}
    \hspace{0.01\textwidth}
    \begin{minipage}{.9\textwidth}
        \subfloat{\includegraphics[width=\textwidth,height=1.37in]{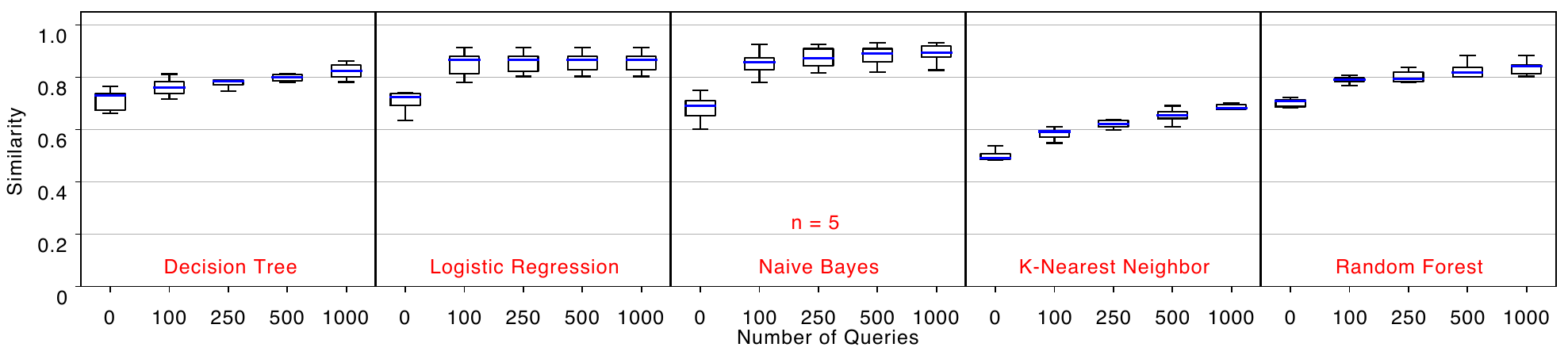}}
    \end{minipage}\\
    \begin{minipage}{.08\textwidth}
        Mushroom Dataset (f)
    \end{minipage}
    \hspace{0.01\textwidth}
    \begin{minipage}{.9\textwidth}
        \subfloat{\includegraphics[width=\textwidth,height=1.37in]{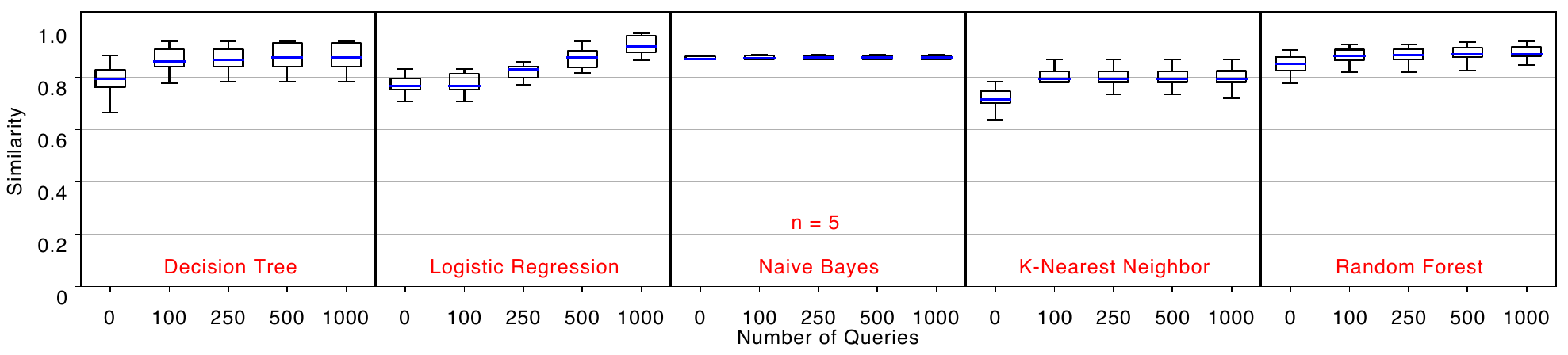}}
    \end{minipage}\\
    
    \caption{Impact of the number of queries ($Q$) on surrogate model similarity when SHAP is used. $(k=3, n=5)$}
    \Description[This figure shows the experimental results of our attack when SHAP is used as the explanation technique.]{This figure shows the impact of number of queries ($Q$) on surrogate model similarity when SHAP is used as the explanation technique. It is a 5 row 6 column figure. Each row represents the dataset used and each column represents the ML model targeted for extraction. Datasets used are Iris (a), Crop (b), Breast Cancer (c), Adult Income (d), Nursery (e), Mushroom (f) from top to bottom. ML models targeted are Decision Tree, Logistic Regression, Naive Bayes, K-Nearest Neighbor and Random forest from left to right. Each figure has an x and y axis. X-axis represents the number of queries sent and the y axis represents the resulting surrogate model's similarity to the target model. In this figure number of features explored (k) is kept at 3 and the auxiliary dataset size per class (n) is kept at 5.}
    \label{fig:similarities2}
\end{figure*}

\subsection{Experiments}
\label{sec:experiments}

We assess the impact of \bl{attack configuration} parameters on the performance of AUTOLYCUS: (i) the query budget ($Q$) which is commonly referred to as the maximum number of queries that can be sent to the MLaaS platform, (ii) the number of samples per class ($n$) in the attacker's auxiliary dataset ($D_A$) and (iii) the method of explanation. Further details are provided in Figure~\ref{fig:similarities1} \bl{and \ref{fig:similarities2}}. 
For each corresponding model type and dataset, we compare the similarity results and the query budget required for AUTLOYCUS to the ones required for the baseline attack, \bl{Steal-ML}, and IWAL attack. \bl{We obtain the results for these attacks from their respective papers. If certain datasets and model types are not explicitly documented or addressed in the papers mentioned above, comparisons are made only when relevant information is available.} There are other significant attack types like EAT~\cite{chandra}, adaptive retraining~\cite{chandra}, Lowd and Meek~\cite{lowd}, and Counterfactuals~\cite{aivodji,wang} \bl{(see Section~\ref{sec:discussion})} to consider in the literature. However, these attacks are incompatible with this study due to the model types (e.g., \bl{non-interpretable models}) they consider.   
Further details about the comparison of our best results between SOTA attacks are provided in Table~\ref{tbl:sota}. 
\bl{It is crucial to highlight that, within our specific attack, LIME explanations offer significantly more information under the current assumptions. This disparity arises from SHAP's provision of feature importances without directional guidance for perturbations. It acts as a semi-blind querying method guided by the order of top features and careful selection of $\delta$. The sole scenario in which attacks using SHAP surpasses attacks using LIME is when local decision boundaries inadequately convey global model behavior.}

For Figures~\ref{fig:similarities1},~\ref{fig:similarities2}, and Table~\ref{tbl:sota}, the number of top features allowed to be explored ($k$) is set to 3. The size of the auxiliary dataset per class ($n$) is set to 1 (for LIME) \bl{and 5 (for SHAP). This is a design choice to demonstrate the impact of the size of the auxiliary dataset while providing a slight leverage to SHAP considering the aforementioned shortcoming. In our additional experiments under similar $n$ settings ($n=1$ for SHAP or $n=5$ for LIME), SHAP performed worse compared to LIME due to having less informativeness. Finally, $\delta$ of SHAP attacks are drawn from the standard deviations and solution space divisions observed in the auxiliary dataset.} 

\noindent\textbf{Iris dataset.} For all considered machine learning models, we observe that the target models reach an accuracy equal to $1$. 
The attacker obtains a surrogate model $S$ that has a similarity \bl{close to} $1$ to the target model $M$ by sending $10$ queries to \bl{most target models} (as shown in Figures~\ref{fig:similarities1} \bl{and \ref{fig:similarities2}(a)). This is attributed to the simplicity of models trained on the Iris dataset. SHAP outperforms LIME due to its higher value of $n$. The proximity of query similarities to the baseline attack when $Q=10$ in SHAP further confirms this.}
Except for naive bayes, when $Q < 50$, AUTOLYCUS significantly outperforms the baseline approach \bl{in both LIME and SHAP attacks}. 
\bl{Compared to the number of queries reported in Table~\ref{tbl:sota} for Steal-ML and IWAL attacks against decision tree and logistic regression models, AUTOLYCUS requires fewer queries, especially as the size of the auxiliary dataset increases.}  

\noindent\textbf{Crop dataset.} 
\bl{Crop dataset contains the highest number of classes ($t$) among the datasets we consider.} The target models reach accuracies above $0.98$ except for the naive bayes model which reaches an accuracy of $0.84$. Due to having high $t$ in the Crop dataset, across all models, the gradual increase of model similarity is more apparent as the number of queries increases. When $Q=1000$, \bl{in Figure~\ref{fig:similarities1}(b) the logistic regression model has a similarity close to $1$. For Figure~\ref{fig:similarities2}(b), decision tree and random forest models have such similarity. 
We make two observations by comparing these results with those obtained for the Iris dataset.  
Firstly, the performance of the model extraction attack is enhanced as the model complexity (architecture, number of features, and classes) decreases or model accuracy increases in accordance with the statement of Rigaki et al.~\cite{rigaki2020survey}. Secondly, even with the extra dataset, LIME sometimes has an edge over SHAP due to the high informativeness of its explanations.}

\noindent\textbf{Breast Cancer dataset.} For the Breast Cancer dataset, target model accuracies are between $0.86$ to $1$. \bl{In Figure~\ref{fig:similarities2}(c), we observe that even a baseline attack can generate models with sufficient accuracy if enough real-life samples are known.} \bl{In Figure~\ref{fig:similarities1}(c),} we observe that the overall similarity between target models and surrogate models consistently increases across different model types as the number of queries increases from $0$ to $100$. Except for decision tree models, with only $100$ queries, there is always at least one surrogate model that achieves close to perfect similarity with target models. 
\bl{When both LIME and SHAP attacks are observed,} the performance of the decision tree model does not increase as $n$ increases from 1 to 5. Because the decision boundaries provided by explanations do not necessarily cover the entire solution space of target models and some classes or branches may remain unexplored. \bl{Our additional experiments on this dataset with higher $k$ provided perfect extractions which can be explained with a more uniform distribution of top important features.} 
Even though the equation solving attack was evaluated in this dataset, it was not provided by Tramer et al.~\cite{10.5555/3241094.3241142} the number of queries needed to extract a logistic regression model trained on this dataset. 
Based on the performances against the Iris dataset, which results in a simpler model extracted with $644$ queries, and the Adult Income dataset, which yields a more complex model extracted with $1485$ queries, we can hypothesize that a \bl{logistic regression} model trained on the Breast Cancer dataset might be extracted with a number of queries somewhere in between. Notably, AUTOLYCUS outperforms the \bl{Equation Solving} attack by achieving extraction with $100$ ($6.4$ times less) queries.

\noindent\textbf{Adult Income dataset.} For the Adult Income dataset in Figures~\ref{fig:similarities1}(d) \bl{and \ref{fig:similarities2}}(d), the target models have accuracies around $0.8$.
The Adult Income dataset consists of more than $100$ features -\bl{when encoded}- and generates complex tree models with a tree depth of more than $20$. \bl{This is a critical dataset that highlights the importance of optimizing all attack configuration parameters for best performance. It requires a higher query limit, a higher number of features considered, a higher and diverse number of auxiliary samples, and an informed set of $\delta$ values provided for perturbation. These dependencies differ between model types.}
For logistic regression, surrogate models with $0.998$ similarity to the target model can be obtained with $1000$ queries by SHAP attacks compared to the $1485$ queries of \bl{Equation Solving} attack. \bl{For decision tree models, $0.937$ similarity can be obtained in $1000$ queries when $n=5$ and $\delta$ set optimized in SHAP attack compared to the $18323$ and $244188$ queries of Path Finding and IWAL attacks. Given better configuration parameters, the performance can be further improved.} 

\noindent\textbf{Nursery dataset.} For the Nursery dataset, the target models have accuracies between $0.95$ to $1$ except for naive bayes. 
This dataset comprises eight \bl{non-binary (when encoded more than 20)} categorical features. \bl{Overall similarity across different models is around $0.91$ for LIME attacks (Figure~\ref{fig:similarities1}(e)) and $0.81$ for SHAP attacks (Figure~\ref{fig:similarities2}(e)).}  
\bl{In this dataset, LIME extractions provide consistently better results compared to SHAP. This is due to the uniform distribution of feature importances and LIME overcomes it with the more informed decision boundaries provided. When $k$ is increased to values closer to $m$ (number of total features), SHAP performs similarly to LIME even when $n$ is as low.}

\noindent\textbf{Mushroom dataset.} Much like the Nursery dataset, the target models of the Mushroom dataset demonstrate accuracies ranging from $0.95$ to $1$, except for naive bayes with accuracy of $0.8$. The Mushroom dataset, along with the Adult Income dataset, boasts the highest number of features (when encoded), resulting in a substantial solution space. In the cases of decision trees and logistic regression, surrogate models with similarities close to or equal to $1$ are achieved with $1000$ queries \bl{in LIME attacks. However, similarities stagnate in SHAP attacks compared to the baseline attack across model types except for logistic regression. Naive Bayes and K-Nearest Neighbor model extractions improve slowly or stagnate compared to other models in both attacks. This indicates that similar to the Nursery dataset, when $k=3$, it does not encompass all the significant features. thus necessitating an increase in $k$. Also in LIME attacks, there is a large variance in the attack's performance for different auxiliary datasets for a fixed $n$ value.   
This is an indicator that the informativeness of auxiliary data can drastically reduce or improve the performance of extraction on these models.}

\begin{figure}[!ht]
    \includegraphics[width=\columnwidth]{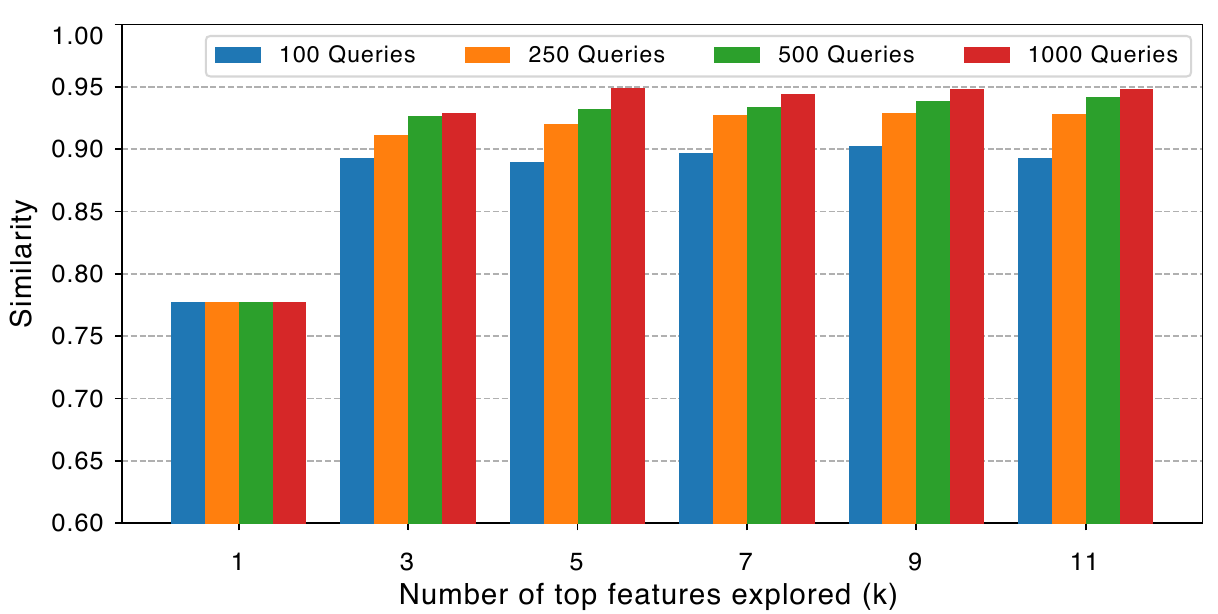}
    \caption{Impact of $k$ on surrogate model similarity.\\$(n=3, LIME)$}
    \Description[This figure explores the impact of top features explored (k) on surrogate model similarity.]{This figure explores the impact of top features explored (k) on surrogate model similarity. These results are obtained from the Adult Income dataset, using LIME as the explainer and the auxiliary dataset size per class (n) as 3. X-axis consists number of features explored (k) from 1 to 11 increasing by 2. }
    \label{fig:features}
\end{figure}
\label{sec:nn}

\noindent\textbf{Number of features explored.} The number of features allowed for traversal ($k$), plays a pivotal role in influencing the performance of AUTOLYCUS. In the experiments, we set $k=3$ (as indicated in Figure~\ref{fig:similarities1} \bl{and \ref{fig:similarities2}}). While this value may be adequate for models trained on datasets with a limited number of features, it might be insufficient for larger models, \bl{especially when a higher number of features are influential in model decision making}. 

We demonstrate the impact of $k$ on a logistic regression model trained on the Adult Income dataset \bl{for a LIME attack. Note that the impact of $k$ on SHAP attacks is similar to the LIME attacks since both attacks utilize feature importances.} Figure~\ref{fig:features} shows the similarity of the surrogate model to the target model for different numbers of queries when varying $k$. 
For $k=1$, the exploration predominantly revolves around a single feature, thereby restricting the surrogate model $S$ to a maximum similarity of $0.77$. With an increase in $k$ up to 5, surrogate models exhibit a top similarity of $0.95$ when $1000$ queries are generated. Beyond $k>4$, the improvement in similarities becomes marginal, suggesting that the target model relies on approximately 5 to 6 crucial features.

%% file: sections/countermeasures.tex
\section{Countermeasures}\label{sec:countermeasures}

\begin{figure*}[!ht]
    \begin{minipage}{.08\textwidth}
        Dynamic \\ Distortion (a)
    \end{minipage}
    \hspace{0.01\textwidth}
    \begin{minipage}{.9\textwidth}
        \subfloat{\includegraphics[width=\textwidth,height=1.3in]{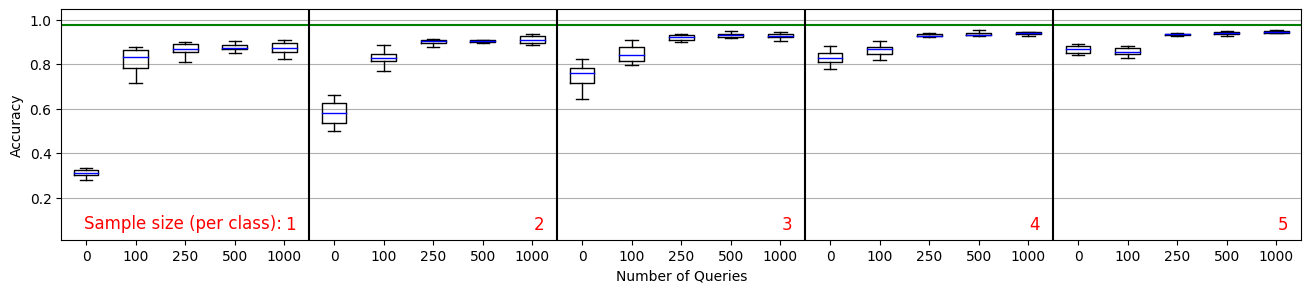}}
    \end{minipage}\\
    \begin{minipage}{.08\textwidth}
        Static \\ Distortion (b)
    \end{minipage}
    \hspace{0.01\textwidth}
    \begin{minipage}{.9\textwidth}
        \subfloat{\includegraphics[width=\textwidth,height=1.3in]{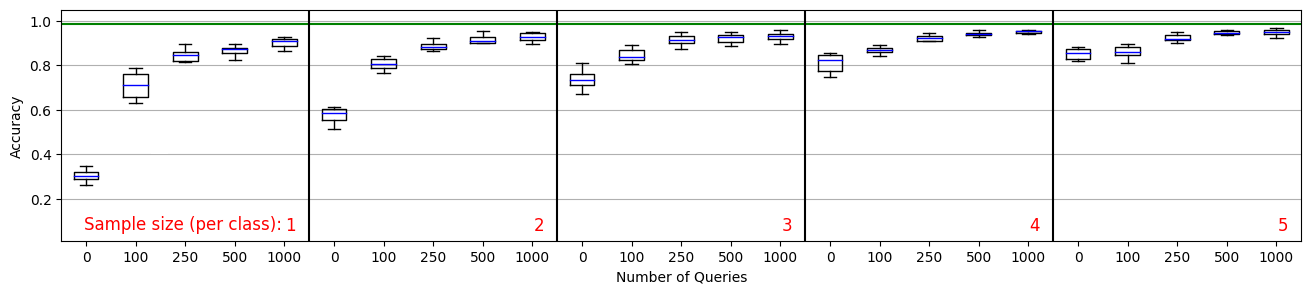}}
    \end{minipage}
    \caption{Impact of countermeasures on the Crop dataset.}
    \Description[This figure demonstrates of the impact of proposed countermeasures on surrogate model similarity.]{This figure demonstrates of the impact of proposed countermeasures on surrogate model similarity. The figure at the top (a) is the dynamic distortion method and the figure at the bottom (b) is the static distortion method. In sub-figures from left to right, auxiliary dataset size per class (n) is increased from 1 to 5. Crop dataset is used for evaluating the performance of proposed countermeasures.}
    \label{fig:countermeasures}
\end{figure*}

In this section, we explore \bl{five} countermeasures that can be employed against AUTOLYCUS. Two of these are based on differential privacy~\cite{dwork}. These methods are influenced by Tramer et al.'s suggestions for prediction minimization~\cite{10.5555/3241094.3241142}.

One of the countermeasures explored in the experiments is differential privacy, which serves as a defense mechanism against the model extraction attack. We examine three types of differentially private methods. The first method involves adding noise to the training dataset of the target model $M$. As noted by Tramer et al., this method does not mitigate the model extraction attacks since the objectives do not align, and applying noise may only impede the learning of sensitive information about the training dataset without targeting the boundaries or parameters of models. We verify this by applying naive differential privacy with calibrated Laplacian noise using $\epsilon$ values between $[1, \ln(3)]$ to the training data. We observe that such application does not mitigate the model extraction attacks and has no significant impact on the decision boundaries. Therefore, as suggested by Tramer et al., we direct our efforts towards targeting the model boundaries.

The second differentially private method we explore is dynamic distortion. It involves applying noise directly to the decision boundaries of \bl{LIME} explanations and changing these boundaries as new queries are sent to the model. 
In LIME, the decision boundaries of each feature are 
static sorted lists, meaning $DB^j=\{db^j1,db^j2,...\}$ per model, where $db_i^j$ is equal to the neighbors of $x_i^j$.\footnote{Neighboring relationship between the states of features ($x_i^j$ values) is based on the corresponding decision boundaries.}  
To implement this countermeasure, we first determine $Q$ queries as the privacy budgets and removed $Q$ samples from the training dataset of the target model. This way for each sample's decision region, we explore how much the decision boundaries of explanations deviate between any neighboring models $M$ and $M'$. Note that these models are trained with $n*t$ and $n*t - Q$ samples, respectively. Considering a sample $X_i$ and for all neighboring models $M$ and $M'$, the maximum difference any decision boundary $d_i^j$ can have can be formalized as follows: $\sigma_i^j = max||M(d_i^j)-M'(d_i^j)||$. 
Then, we calculate a Laplacian distribution with deviation corresponding to $\sigma$ as the sensitivity, and for varying $\epsilon$ values, we apply Laplacian noise directly to explanation decision boundaries such that they are changed in every query. 
The results for dynamic distortion when $\epsilon=1$ are shown in Figures~\ref{fig:countermeasures} (a). This method not only fails to mitigate the model extraction attack, but it also enhances the attack strength by allowing a more diverse and informative surrogate dataset to be explored.

The third method is static distortion. It's a generic random noise application, using the dynamic distortion method with $\epsilon=1$. In this method though, the noise drawn from the Laplace distribution is unchanged across queries such that the model explanations seem like they are drawn from another model. The results for static distortion are shown in Figure~\ref{fig:countermeasures} (b). We observe that this method also fails to mitigate the model extraction attack and, like the previous method, makes the attack even stronger by improving the accuracy and similarity of resulting surrogate models. However, the improvement is negligible compared to the second method.

Applying noise with sufficiently low $\epsilon$ values can distort (dynamic and static) the explanation boundaries to prevent model extraction attacks. However, the utility of model explanations is hindered more than the model extraction success, in such instances. Experiments with lower $\epsilon$ values, produced mathematically impossible decision boundaries due to excessive perturbation. 

\bl{The fourth method involves adding noise to the feature importances from SHAP. These importances dictate the sequence of feature perturbations based on an attacker-defined $\delta_j$ value. While significant noise could disrupt this sequence and cause deviations in near to $k^{th}$ feature perturbations, unless the value of $k$ is sufficiently low, these deviations would normalize with increasing queries. However, this normalization would hinder the utility of explanations, potentially misguiding users about their predictions. Thus, using noise on SHAP explanations is unlikely to effectively counter model extraction attacks.}

Another countermeasure suggested by Tramer et al. is rounding the confidence values obtained during predictions. This method is intuitively effective against models with confidence values and extraction attacks that deterministically solve boundaries (i.e., equation-solving attacks, path-finding attacks). Since AUTOLYCUS reflects the decision boundaries in the re-training process stochastically and only employs labels and the decision boundaries of explanations, methods such as rounding confidences or explanation strength do not have any impact on its performance.

Prediction minimization methods are prone to failure in AUTOLYCUS, but countermeasures like setting query budgets can slow down the extraction process. While not explored in this study, fingerprinting against malicious surrogate model sharing~\cite{lukas} is a potential future work.

%% file: sections/discussion.tex
\section{Discussion}
\label{sec:discussion}

\begin{figure*}[!ht]
    \centering
    \begin{minipage}{0.1\textwidth}
        LIME MLP \\ Extraction (a) \\ (k=3, n=1)
    \end{minipage}
    \hspace{0.02\textwidth}
    \begin{minipage}{.8\textwidth}
        \subfloat{\includegraphics[width=\textwidth,height=1.3in]{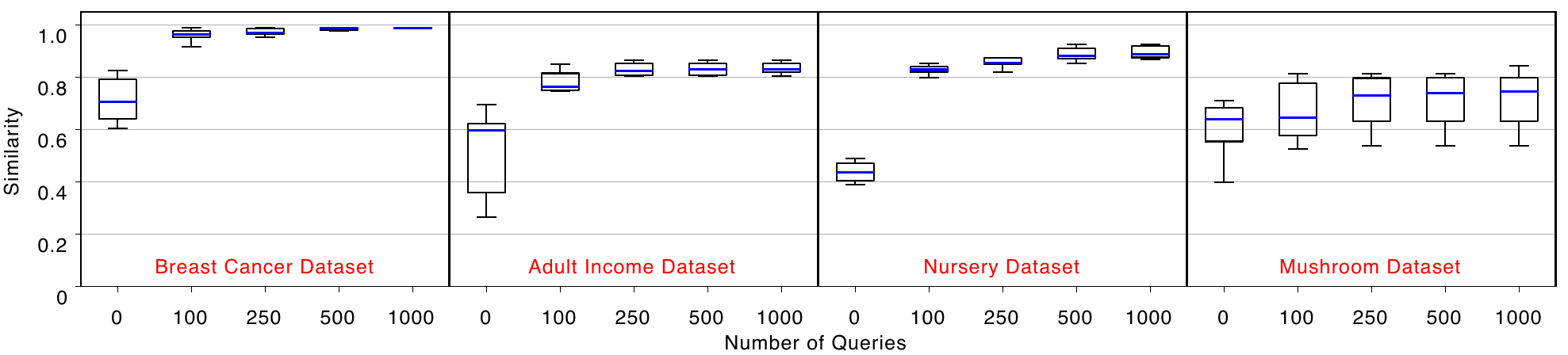}}
    \end{minipage}\\
    \begin{minipage}{.1\textwidth}
        SHAP MLP \\ Extraction (b) \\ (k=3, n=5)
    \end{minipage}
    \hspace{0.02\textwidth}
    \begin{minipage}{.8\textwidth}
        \subfloat{\includegraphics[width=\textwidth,height=1.3in]{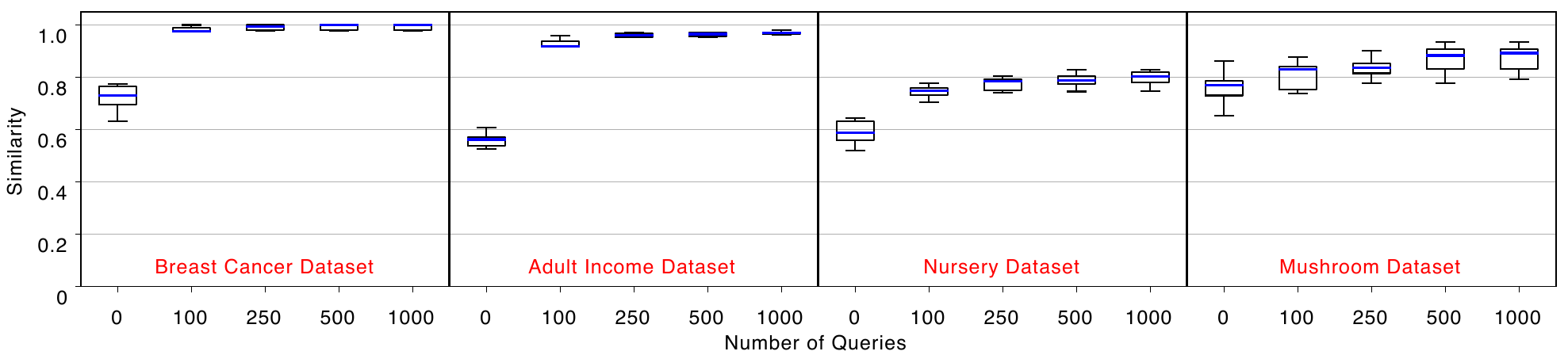}}
    \end{minipage}
    \caption{Multilayer perceptron surrogate model similarities across different datasets.}
    \Description[This figure explores the performance of AUTOLYCUS on non-interpretable machine learning models.]{This figure explores the performance of AUTOLYCUS on non-interpretable machine learning models. Multilayer perceptrons are used for evaluation. The top figure shows the results with LIME as the explainer, whereas the bottom figure shows the results with SHAP as the explainer. In both figures, the number of features explored (k) is set to 3. However, auxiliary dataset size per class is set to 1 in LIME and 5 in SHAP. From left to right, Breast Cancer, Adult Income, Nursery and Mushroom datasets are used in evaluation.}
    \label{fig:mlp}
\end{figure*}

\noindent\textbf{Limitations.} The primary limitation of AUTOLYCUS is the probabilistic reconstruction of models in comparison to the exact reconstruction by the \bl{white-box} attacks \bl{(Equation Solving, Path-finding)} of Tramer et al.~\cite{10.5555/3241094.3241142}. AUTOLYCUS involves retraining models in a black-box setting, and therefore \bl{attack configuration} parameters such as explanations, traversal, or auxiliary datasets are subject to limitations in some models.

For instance, outlier classifications that are not adequately represented in explanations or auxiliary datasets (e.g., edge cases in decision tree models) may remain unexplored during the traversal procedure. This limitation is particularly evident in complex models with a high number of top features, where some branches may not be properly represented in the surrogate model, reducing their similarity to the target model. An example limitation is observed in the experiments \bl{where LIME is used for targeting} a decision tree model trained on the Adult Income dataset. Despite producing surrogate models with similar accuracy, the low accuracy and multiple influential features of this model made it difficult to adequately discover and functionally represent certain aspects of the model with the assumptions and small auxiliary datasets available.

Furthermore, AUTOLYCUS's effectiveness is limited to the type of data that target models are trained on. AUTOLYCUS performs well on tabular and mixed data suitable for interpretable models and decision boundaries. However, in image classification tasks, \bl{tabular LIME and SHAP} explanations of single pixels have equivalent strengths, making these individually considered features and their boundaries uninformative, effectively rendering the attack useless. Since image classifications consider a collection of pixels or regions in figures, explanations, and traversal suited for such a task must be employed. Therefore, AUTOLYCUS should be employed against models with informative explanations. 

\bl{In contrast with evaluations in Section~\ref{sec:experiments}, the performance of AUTOLYCUS is influenced by careful adjustment of the following attack configuration and model parameters: (i) ensuring sufficiently high number of queries are sent; (ii) selecting adequate number of features ($k$) for perturbation, guided by the distribution of feature importance; (iii) incorporating sufficient number of (iv) informative auxiliary samples; (v) considering complexity of the models -lower complexity models are extracted easier-, including any unknowns or parameters in neural network terminology; (vi) ensuring the accuracy of the model, (vii) optimizing selection of $\delta$ values especially when SHAP is used as the XAI tool. These factors collectively contribute to the efficacy of AUTOLYCUS in adversarial settings.}

\bl{\noindent\textbf{Other model-agnostic explainability tools.} \bl{In the realm of alternative explanation techniques, the integration of counterfactuals offers an additional source of valuable information, with various documented attacks in the literature~\cite{aivodji,wang}. Our goal with LIME and SHAP explanations is similar to replicating the utility of counterfactuals with additional steps. While providing counterfactuals directly would streamline our framework, it would also provide limited novelty. Moreover, due to documented privacy risks, MLaaS providers may refrain from using these explanations in commercial settings. Due to these concerns, we have opted not to explicitly include them in our framework.}}

\noindent\textbf{Application to other machine learning models.} 
After obtaining promising results against interpretable models, we investigate the effectiveness and transferability of AUTOLYCUS against black-box models. Our initial focus lies on neural networks. We conduct experiments on four datasets, with a configuration similar to the \bl{Steal-ML} attack, where target models consist of multilayer perceptrons with a single hidden layer comprising 20 nodes. \bl{Only for the Breast Cancer dataset, hidden layers are increased to 100 nodes due to the model's low accuracy of $0.22$ on 20 nodes.} 
We set $k=3$ and $n=1$ (for LIME) and $n=5$ (for SHAP) following the setting in Section~\ref{sec:experiments}. Figure~\ref{fig:mlp} shows the similarities for each dataset.

For the Breast Cancer dataset, the attacker obtains surrogate models exhibiting a similarity of up to $1$ with $100$ queries \bl{in both LIME and SHAP attacks}. This outcome aligns with our expectations given the dataset's relatively low complexity, as demonstrated by the results shown in Figures~\ref{fig:similarities1} \bl{and \ref{fig:similarities2}}(c).
Furthermore, for both Adult Income and Nursery datasets, surrogate models achieve comparable performances when the target models are random forests \bl{and LIME is used}. \bl{On the other hand, while the SHAP attack performs better than LIME attacks due to increased auxiliary dataset size, it is less efficient in extracting the neural network trained on the Nursery dataset compared to LIME attacks. This discrepancy is attributed to differences in explanation informativeness. LIME excels in explaining categorical solution spaces with fewer unknowns when $k$ is low. As the solution space of the Nursery dataset is relatively small compared to Adult Income and Mushroom datasets, LIME explanations retain their efficacy against SHAP explanations even with a low value of $n$. However, with an increase in $k$, the performance of the SHAP attack begins to improve, similar to the behavior observed with LIME explanations.}
In the Mushroom dataset, an increased number of queries does not yield a desired improvement in the similarity of surrogate models \bl{for LIME attack.} Experiments involving higher values of $k$ also do not yield significant performance enhancement observed in the case of interpretable models. This limitation may be attributed to the scarcity of queries required to accurately represent the \bl{complex} boundaries of neural networks \bl{and the number of features with high solution space further complicating the extraction on relatively low query budgets. In the SHAP attack, extraction is more successful owing to the larger number of available auxiliary samples ($n$), and this success tends to increase with higher numbers of queries. This suggests that a greater abundance of samples facilitates the exploration of rarer cases (which seems to be more abundant in the Mushroom dataset) more effectively.} 
Additionally, in the process of extracting the complex boundaries of neural networks, LIME \bl{and SHAP} explanations may exhibit occasional inconsistencies, which could constitute an additional factor, hindering performance. \bl{In conclusion, under comparable conditions with sufficiently large values of $n$ and $k$, LIME mostly demonstrates superior performance in capturing similarities within neural networks compared to SHAP.}

%% file: sections/conclusion.tex
\section{Conclusion}
\label{sec:conclusion}
In this paper, we have proposed AUTOLYCUS, a novel model-agnostic extraction attack that exploits model-agnostic AI explanations to extract the decision boundaries of interpretable models. 
The empirical results on six different datasets have shown the efficacy of the attack. 
In particular, we have demonstrated that an attacker can exploit AI explanations to create fairly accurate surrogate models that have high similarity to the target models even under low query budgets. 
\bl{We have also demonstrated that AUTOLYCUS requires fewer queries for partial reconstructions with comparable accuracy and similarity than the state-of-the-art attacks that rely on exact reconstruction.}
Furthermore, we have explored potential countermeasures to mitigate this attack such as adding noise to the decision boundaries of explanations or adding noise to the training data.
The empirical results have shown that these countermeasures are ineffective against AUTOLYCUS. 
As a future work, we plan to study the efficacy of multiple model explanations, potential increased risks due to explanations in black-box machine learning models such as convolutional neural networks, and work on new and effective countermeasures.

%% file: sections/appendix.tex
\appendix
\newpage
\section{XAI in MLaaS Platforms}
\label{sec:appendix}
\bl{Major MLaaS platforms such as Azure Machine Learning (Responsible AI)~\cite{azureExp}, Amazon SageMaker (Clarify)~\cite{awsClarify}, Watson Machine Learning (OpenScale)~\cite{ibmExp} and Google Cloud (Vertex Explainable AI)~\cite{googleCloud} offer various explanation techniques. Some of the provided techniques in these platforms are included but not limited to SHAP~\cite{shap}, LIME~\cite{10.1145/2939672.2939778}, Permutation Feature Importance (PFI)~\cite{pfi}, Partial Dependence Plots (PDP)~\cite{pdp}, Integrated Gradients~\cite{gradients} and XRAI~\cite{xrai}.} However most explainers offered by MLaaS providers \cite{googleCloud,azureExp,awsClarify} are predominantly white-box, intended primarily for model developers. They provide detailed insights into the inner workings of the models, contrasting with the simpler, black-box interfaces often provided to the querying users. These use cases include, but are not limited to, detecting bias, assessing fairness, detecting outliers, and dimensionality reduction. Consequently, for commercialized black-box models, explainers should provide utility for a given sample but don't reveal the entire inner workings of the models. Thus, explainers employed must be local, preferably model-agnostic, and post-hoc. The family of explainers that satisfy these criteria for the models we consider is limited. It consists of SHAP, LIME, counterfactuals/what-ifs (which are discussed in Section~\ref{sec:discussion}) and feature importances (already returned by LIME and SHAP).